\pdfoutput=1

\documentclass[11pt]{article}
\usepackage{acl}

\usepackage{times}
\usepackage{latexsym}

\usepackage[T1]{fontenc}

\usepackage[utf8]{inputenc}

\usepackage{microtype}

\usepackage{inconsolata}

\usepackage{amsmath,amsfonts,bm}

\def\eqref#1{equation~\ref{#1}}

\def\1{\bm{1}}

\DeclareMathAlphabet{\mathsfit}{\encodingdefault}{\sfdefault}{m}{sl}
\SetMathAlphabet{\mathsfit}{bold}{\encodingdefault}{\sfdefault}{bx}{n}

\usepackage{booktabs}
\usepackage{siunitx}
\usepackage{multirow}
\usepackage{amsfonts}
\usepackage{amsmath}
\usepackage{amssymb}
\usepackage{xspace}
\usepackage{booktabs}
\usepackage{graphicx}
\usepackage{floatrow}
\usepackage{comment}
\usepackage{float}
\usepackage{listings}
\usepackage{changepage}
\usepackage[flushleft]{threeparttable}

\usepackage{tikz}
\usepackage{circledsteps}
\usepackage{adjustbox}
\usepackage[T1]{fontenc}
\usepackage{textcomp}

\usepackage{cleveref}

\usepackage{bm}
\usepackage{fix-cm}
\usepackage{amssymb}
\usepackage{amsthm}

\usepackage{enumerate}

\usepackage[disable]{todonotes}

\usepackage[most]{tcolorbox}

\newtcolorbox{promptbox}[2][]{%
    breakable,
    colback=gray!5,
    colframe=white,    
    opacityframe=0,    
    coltitle=white,    
    colbacktitle=gray!90!black, 
    title=#2,
    fonttitle=\bfseries,
    halign=flush left,
    before title=\vspace*{1.5mm}, 
    after title=\vspace*{1.5mm},  
    #1
}

\newtcolorbox{systemmessagebox}{
    breakable,
    colback=red!5, 
    colframe=red!75!black, 
    title=System prompt,
    fonttitle=\bfseries
}

\newtcolorbox{fewshotbox}{
    breakable,
    colback=gray!15, 
    colframe=gray!75!black, 
    title=Few-shot examples (optional),
    fonttitle=\bfseries
}

\newtcolorbox{usermessagebox}{
    breakable,
    colback=green!5, 
    colframe=green!75!black, 
    title=User,
    fonttitle=\bfseries
}

\newtcolorbox{assistantmessagebox}{
    breakable,
    colback=blue!5, 
    colframe=blue!75!black, 
    title=Assistant,
    fonttitle=\bfseries
}

\makeatletter
\@ifpackageloaded{lineno}{
    \newenvironment{prompt}[1][Prompt]{%
        \nolinenumbers
        \small%
        \promptbox{#1}
    }{%
        \endpromptbox
        \linenumbers  
    }
}{
    \newenvironment{prompt}[1][Prompt]{%
        \small%
        \promptbox{#1}
    }{%
        \endpromptbox
    }
}
\makeatother

\newcommand{\messageseparator}{%
\textcolor{lightgray}{\rule{\linewidth}{0.5pt}}\vspace{0.2cm}%
}

\newcommand{\stepseparator}{%
\textcolor{lightgray}{\rule{0.8\linewidth}{0.25pt}}\vspace{0.1cm}%
}

\newenvironment{systemmessage}{\textbf{System prompt:}\\}{

}

\newenvironment{smalljinja}{%
    \small%
    \let\oldjinja\jinja%
    \renewcommand{\jinja}[1]{\oldjinja{##1}}%
}{%
    \let\jinja\oldjinja%
}

\newtcolorbox{describedexamplebox}[1][]{%
    colback=gray!15,
    colframe=white,    
    opacityframe=0,    
    coltitle=white,    
    colbacktitle=gray!90!black, 
    fonttitle=\bfseries,
    halign=flush left,
    before title=\vspace*{1.5mm}, 
    after title=\vspace*{1.5mm},  
    boxsep=5pt,left=0pt,right=0pt,top=0pt,bottom=0pt
    #1
}

\definecolor{dimgray}{rgb}{0.41, 0.41, 0.41}

\usepackage{etoolbox} 

\newbool{firstelement}
\global\setbool{firstelement}{false}

\newenvironment{fewshotexamples}{%
\messageseparator

\textbf{Few shot examples:\vspace{0.2cm}}%
\global\setbool{firstelement}{true}
\begin{adjustwidth}{0.4cm}{}
}{\end{adjustwidth}%
\global\setbool{firstelement}{false}

}

\usepackage{adjustbox}

\newlength{\iconwidth}
\newlength{\iconpadding}
\newenvironment{usermessage}
  {\ifbool{firstelement}{%
    \global\setbool{firstelement}{false} 
  }{\messageseparator}%
  \par\noindent
   \adjustbox{valign=b}{\includegraphics[width=\iconwidth]{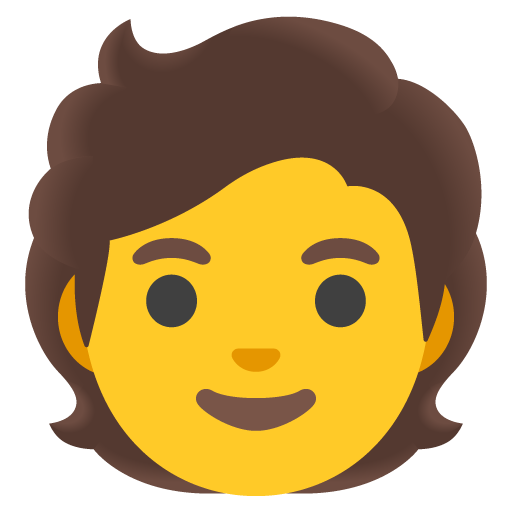}}
   \hspace{\iconpadding}\textbf{User:}
   \vspace{2mm}
   \begin{adjustwidth}{\iconwidth+\iconpadding}{}
   }
  {\end{adjustwidth}
  
  }

\newenvironment{task_description}{%

\textbf{Task description:\vspace{0.2cm}}%
\global\setbool{firstelement}{true}
\begin{adjustwidth}{0.4cm}{}
}{\end{adjustwidth}%
\global\setbool{firstelement}{false}

}

\newenvironment{nav_instruction}{%
\messageseparator

\textbf{Navigation instruction:\vspace{0.2cm}}%
\global\setbool{firstelement}{true}
\begin{adjustwidth}{0.4cm}{}
}{\end{adjustwidth}%
\global\setbool{firstelement}{false}

}

\newenvironment{nav_instruction_nl}{%

\textbf{Navigation instruction:\vspace{0.2cm}}%
\global\setbool{firstelement}{true}
\begin{adjustwidth}{0.4cm}{}
}{\end{adjustwidth}%
\global\setbool{firstelement}{false}

}

\newenvironment{trajectory}{%
\messageseparator

\textbf{Navigation trajectory:\vspace{0.2cm}}%
\global\setbool{firstelement}{true}
\begin{adjustwidth}{0.4cm}{}
}{\end{adjustwidth}%
\global\setbool{firstelement}{false}

}

\newenvironment{obs}{%
\tt \vspace{2mm}

\par\noindent
   \adjustbox{valign=c}{(\includegraphics[width=\iconwidth]{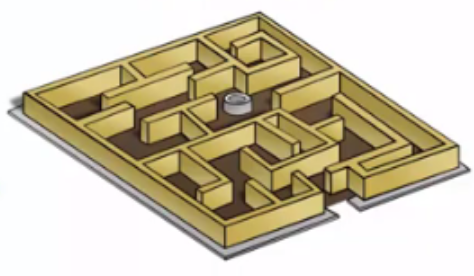}}
   \hspace{\iconpadding}\textbf{environment:)}

\global\setbool{firstelement}{true}
\begin{adjustwidth}{0.4cm}{}
}{\end{adjustwidth}%
\global\setbool{firstelement}{false}
\vspace{2mm}
}

\newenvironment{ans}{%
\tt \vspace{2mm}

\par\noindent
   \adjustbox{valign=c}{(\includegraphics[width=\iconwidth]{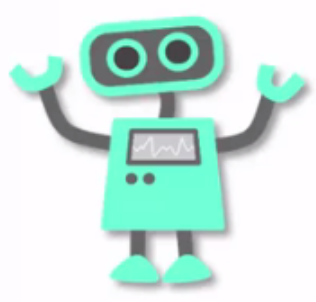}}
   \hspace{\iconpadding}\textbf{navigation agent:)}

\global\setbool{firstelement}{true}
\begin{adjustwidth}{0.4cm}{}
}{\end{adjustwidth}%
\global\setbool{firstelement}{false}
\stepseparator
\vspace{2mm}
}

\newenvironment{ans_nl}{%
\tt \vspace{2mm}

\par\noindent
   \adjustbox{valign=c}{(\includegraphics[width=\iconwidth]{emojis/robot.png}}
   \hspace{\iconpadding}\textbf{navigation agent:)}

\global\setbool{firstelement}{true}
\begin{adjustwidth}{0.4cm}{}
}{\end{adjustwidth}%
\global\setbool{firstelement}{false}
\vspace{2mm}
}

\lstdefinestyle{jsonstyle}{
  basicstyle=\ttfamily,
  showstringspaces=false,
  breaklines=true,
  frame=none,
  upquote=true,
  literate=
    {:}{{{\color{purple}{:}}}}{1}
    {,}{{{\color{purple}{,}}}}{1}
    {\{}{{{\color{brown}{\{}}}}{1}
    {\}}{{{\color{brown}{\}}}}}{1}
    {[}{{{\color{brown}{[}}}}{1}
    {"}{{"}}{1} 
    {]}{{{\color{brown}{]}}}}{1},
}

\lstnewenvironment{jsonblock}[1][]%
  {\lstset{style=jsonstyle, language=Java, #1}}%
  {}

\newcommand{\jinja}[1]{\lstinline[style=jsonstyle,keepspaces=true]|\{\{ #1 \}\}|}

\usepackage{times}
\usepackage{latexsym}

\usepackage[T1]{fontenc}

\usepackage[utf8]{inputenc}

\usepackage{microtype}

\usepackage{inconsolata}

\usepackage[utf8]{inputenc} 
\usepackage[T1]{fontenc}    
\usepackage{hyperref}       
\usepackage{xspace}
\usepackage{amsmath}
\usepackage{cleveref}
\crefname{section}{§}{§§}
\Crefname{section}{§}{§§}
\crefformat{subsubsection}{\S#2#1#3}
\usepackage{url}            
\usepackage{booktabs}       
\usepackage{amsfonts}       
\usepackage{nicefrac}       
\usepackage{microtype}      
\usepackage{xcolor}         
\usepackage{colortbl}
\usepackage{multirow}

\usepackage{appendix}
\usepackage{algpseudocode,algorithm,algorithmicx}
\usepackage{bm}
\usepackage{mathtools}
\usepackage{makecell}
\usepackage{enumitem}
\usepackage[labelfont=bf]{caption}
\usepackage{subcaption}
\usepackage{lipsum}
\usepackage{wrapfig}
\usepackage{cuted}
\usepackage{xcolor}
\usepackage{adjustbox}
\usepackage{multirow}
\usepackage{graphicx}
\usepackage{gensymb}
\usepackage{listings}
\usepackage{floatrow}

\newcolumntype{H}{>{\setbox0=\hbox\bgroup}c<{\egroup}@{}}
\definecolor{cadmiumgreen}{rgb}{0.0, 0.52, 0.24}
\definecolor{mediumpersianblue}{rgb}{0.2, 0.4, 0.8}
\usepackage{pifont}
\definecolor{darkgreen}{rgb}{0,0.6,0}
\definecolor{codegray}{rgb}{0.5,0.5,0.5}
\definecolor{lightgray}{rgb}{0.7,0.7,0.7}
\definecolor{codepurple}{rgb}{0.07,0,0.53}
\definecolor{codered}{RGB}{189,41,0}
\definecolor{codecomment}{RGB}{153,153,153}
\definecolor{backcolour}{rgb}{0.96,0.96,0.96}
\definecolor{mygreen}{rgb}{0.0, 0.5, 0.0}
\definecolor{royalblue}{rgb}{0.0, 0.14, 0.4}
\definecolor{egyptianblue}{rgb}{0.06, 0.2, 0.65}
\definecolor{royalazure}{rgb}{0.0, 0.22, 0.66}
\definecolor{portlandorange}{rgb}{1.0, 0.35, 0.21}
\definecolor{saddlebrown}{RGB}{139,69,19}
\definecolor{sienna}{RGB}{183,105,68}
\definecolor{saddlebrown}{RGB}{139,69,19}
\hypersetup{
    colorlinks=true,
    linkcolor=sienna,
     urlcolor=royalblue,
    citecolor=egyptianblue,
}

\title{LangNav: Language as a Perceptual Representation for Navigation}
\author{Bowen Pan$^\diamond$ \hspace{5mm} Rameswar Panda$^\dagger$ \hspace{5mm}  SouYoung Jin$^\star$ \hspace{5mm} Rogerio Feris$^\dagger$ 
\\
\textbf{Aude Oliva$^{\diamond\dagger}$ \hspace{5mm} Phillip Isola$^\diamond$ \hspace{5mm} Yoon Kim$^\diamond$}
\\
$^\diamond$MIT CSAIL, $^\dagger$MIT-IBM Watson AI Lab, $^\star$Dartmouth College
\\
\small{\texttt{\{bpan, oliva, phillipi, yoonkim\}@mit.edu,}} \\
\small{\texttt{rpanda@ibm.com,}}
\small{\texttt{rsferis@us.ibm.com,}}
\small{\texttt{souyoung.jin@dartmouth.edu}}
}

\newcommand{\ours}{LangNav\xspace}

\begin{document}

\maketitle
\vspace{-2mm}
\begin{abstract}
\vspace{-1.5mm}
We explore the use of language as a perceptual representation for vision-and-language navigation (VLN), with a focus on low-data settings. Our approach uses off-the-shelf vision systems  for image captioning and object detection to convert an  agent's egocentric panoramic view at each time step  into natural language descriptions. We then finetune a pretrained language model to select an action, based on the current view and the  trajectory history, that would best fulfill the navigation instructions. In contrast to the standard setup which adapts a pretrained language model to work directly with continuous visual features from pretrained vision models, our approach instead uses (discrete) language as the perceptual representation. We explore several use cases of our language-based navigation (\ours) approach on the R2R VLN benchmark: generating synthetic trajectories from a prompted  language model (GPT-4) with which to finetune a smaller language model; domain transfer where we transfer a policy learned on one simulated environment (ALFRED) to another  (more realistic) environment (R2R); and combining both vision- and language-based representations for VLN. Our approach is found to improve upon  baselines that rely on visual features in settings where only a few expert trajectories (10-100) are available, demonstrating the potential of language as a perceptual representation for navigation.
\end{abstract}

\section{Introduction}
Applications of large language models (LMs) to non-linguistic embodied tasks have generally focused on using the implicit world knowledge within LMs to predict sub-tasks and actions for planning \citep{ahn2022can,huang2022inner,huang2022language,singh2022progprompt}. 
For instance, recent work has shown that LMs can be prompted to create a list of actions (e.g.,  \texttt{GoToBathroom}, \texttt{LocateToothbrush}) given a high-level goal given in natural language (e.g., ``brush teeth'')  \citep{huang2022language}. These approaches rely on the LM's  priors on action sequences and inter-object correlations  acquired through large-scale pretraining  \citep{zhou2023esc,li2023lampp,zhao2023large}, and it has not been clear whether   text-only models can be  finetuned for  tasks such as vision-and-language navigation which requires an egocentric agent  follow   instructions to navigate a 3D environment using  visual input.

To be clear, there \emph{is} a substantial body of work on using pretrained LMs for vision-and-language navigation tasks \citep[][\textit{inter alia}]{Hong_2021_CVPR,qi2021road,qiao2022hop}. The standard approach is to use a pretrained LM over the natural language instructions to extract text features that are combined with the agent's perceptual representations, which are given by continuous image features extracted from pretrained vision models \citep{wang2019reinforced, hao2020prevalent}. While effective in data-rich regimes, the direct use of vision features makes the approach difficult to apply in cases where only a few  labeled trajectories exist (e.g., 10 trajectories), as these approaches need to learn a full joint vision-language module that combines a pretrained vision model with a pretrained text model. A popular strategy in such low data regimes is to generate synthetic data or transfer knowledge from other domains. However, generating realistic perception data is itself a difficult task, and domain transfer with models that  rely purely on visual features can overfit to the non-transferable features \citep{anderson2021sim}. 
\begin{figure*}[t]
\centering
\includegraphics[width=0.95\linewidth]{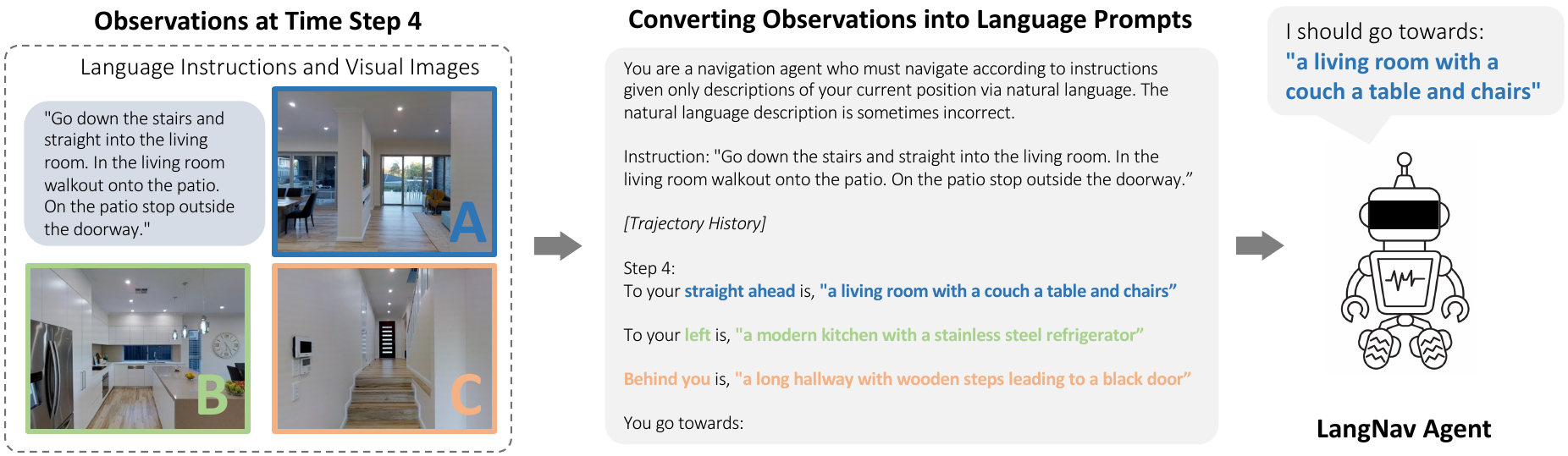}
\caption{Overview of language-based navigation (\ours). We describe the task instructions and visual observations (from off-the-shelf vision systems) through text. A language model is then finetuned  to predict which direction to move towards based on the language descriptions. Here, views \textcolor{mediumpersianblue}{\textbf{A}}, \textcolor{darkgreen}{\textbf{B}}, and \textcolor{pink}{\textbf{C}} correspond to the \textcolor{mediumpersianblue}{front}, \textcolor{darkgreen}{left}, and \textcolor{pink}{rear} views of the agent.}
\label{fig:teaser}
\end{figure*}

This paper explores an alternative approach for  vision-and-language navigation  by exploiting language itself as the perceptual representation space. Our approach uses off-the-shelf vision models to obtain textual descriptions of the agent's egocentric panoramic view. The text descriptions are then fed to an LM which must select the next action given the instruction and (text descriptions of) the previous actions or observations. See Figure~\ref{fig:teaser} for an overview.
The use of language to represent an agent's perceptual field makes it possible to readily utilize the myriad capabilities of  language models, especially when the training data is limited. In our first case study, we show how we can use a small amount of seed training data (10-100 trajectories) to cheaply obtain synthetic ``trajectories'' from a powerful but closed-source LM \cite[GPT-4;][]{openai2023gpt4}. We find that finetuning a smaller language model \cite[LLaMA/LLaMA2; ][]{touvron2023llama,touvron2023llama2} on the generated trajectories mixed with the original seed data results in a langauge-based navigation agent that  outperforms a vision-based agent that is finetuned on the same seed data. In our second study, we explore the use of language as a domain-invariant  representation to perform domain transfer, where we transfer an agent trained on a computer-generated environment \citep[ALFRED;][]{ALFRED20} to the real-world R2R \citep{anderson2018vision} environment. Insofar as language is hypothesized to have co-evolved with the human brain to enable efficient communication \citep{deacon1997symbolic}, it naturally abstracts away low-level perceptual details, and we indeed find that \ours exhibits improved transfer compared to the vision-based agent. We further show that language can provide further benefits even in the presence of vision-based features.  
Our results collectively suggest that language as a perceptual representation can be helpful in the low-data navigation settings.

\vspace{-2mm}
\section{Background: Room-to-Room Vision-language Navigation}\label{sec:method:overview}
\vspace{-2mm}
A popular testbed for  vision-and-language navigation (VLN) is the room-to-room dataset \citep[R2R; ][]{anderson2018vision}, in which an agent must perceive and navigate a real-world 3D environment based on a language instruction $U$ and an initial state $S_0$. At each time step $t$, the agent uses the current observation $O_t$, the original language instructions $U$, and the trajectory history $H_t$, to predict the panoramic action $a_t$. The current observation is given by a set of panoramic images that describe the agent's egocentric view, i.e., $O_t = \{ I_{t,0}, ..., I_{t,V}\}$ where $V$ corresponds to the number of discretized view angles.\footnote{In R2R this can be as many as 36 (12 headings and 3 elevations). However we follow previous works only consider the navigable views, which is often many fewer than 36.}  The panoramic action $a_t$ corresponds to which navigable view in $O_t$ to go towards, i.e., $a_t \in O_t$. After selecting an action, the state transitions from $S_t$ to $S_{t+1}$. The aim is to output the command \texttt{STOP} after reaching the goal $G$ specified by $U$ in state $S_0$.

The standard approach in R2R is to process the panoramic images $\{ I_{t,0}, ..., I_{t,V}\}$ with a pretrained visual encoder $E_v$ to extract continuous visual features $F_{t,v} = \{ E_v(I_{t,0}), ..., E(I_{t,V})\}$. The language instruction is typically processed by a pretrained language encoder $E_l$ (e.g., BERT \cite{devlin-etal-2019-bert}) to extract the language features $F_{l} = E_l(U)$. These features, along with a hidden state representation of the trajectory history $h_{t-1}$, are fed to a joint vision-language module (e.g., another Transformer) that attends over $\{ I_{t,0}, ..., I_{t,V}\}$ to select the action $a_t$.

\vspace{-2mm}
\section{Language as a Perceptual Representation for Navigation}\label{sec:method:lan}
\vspace{-2mm}
We begin by describing the perception-to-text models employed for converting  visual observations into text (\cref{sec:vision-text}). We then discuss the prompt templates for converting the text into natural language (\cref{sec:prompt}),  followed by a description of the offline imitation learning algorithm for learning (\cref{sec:bc}). 

\vspace{-1mm}
\subsection{Vision-to-text System} \label{sec:vision-text}
\vspace{-1mm}
We use off-the-shelf vision models to convert visual observations into language descriptions. 
Specifically, we use an image captioning model \cite[BLIP;][]{li2022blip} and an object detection model \cite[Deformable DETR;][]{zhu2020deformable} over each view angle $I_{t,j}$ to obtain the text descriptions,
\begin{align*}
    &C_{t, j} = \textsc{ImageCaptioner} (I_{t,j}),  \\
    &x_{t,j,0}, \dots, x_{t,j,M} = \textsc{ObjectDetector}(I_{t,j}),
\end{align*}
where $M$ is the number of detected objects.\footnote{We did not experiment  much with different off-the-shelf vision systems and quickly converged on these  two models which seemed to  produce reasonable results. Since \ours separates perception from navigation, we expect that advances made in perception (e.g., through better captioning systems) will automatically result in improvements to our system, which is a nontrivial advantage of our approach compared to systems that entangle perception and navigation into a single model.} 
 
\vspace{-1mm}
\subsection{Prompt Templates}\label{sec:prompt}
\vspace{-1mm}
Figure~\ref{fig:teaser} illustrates how the image caption and the detected objects are combined via templates to construct pieces of text on which to condition the  language model. Based on the prompt template, the language model will be finetuned on the (language representations of)  output actions $\{a_1, \dots, a_T\}$. We briefly describe the prompt template (see \cref{sec:app-full} for a full example).

\vspace{-1mm}
\paragraph{Task description $D$.} The task description is given by: \vspace{-1mm}
\begin{quote}
\footnotesize \texttt{You are a navigation agent who must navigate according to instructions given only descriptions of your current [...]}. \vspace{-2mm}
\end{quote}
\vspace{-3mm}
\paragraph{Navigation instruction $U$.} The navigation instruction, which provides instructions to the agent on how to reach the goal, can be from  R2R  (our main dataset), synthesized by GPT-4 (for data augmentation), or  ALFRED  (for domain transfer). An example instruction from  R2R is: 
\begin{quote}
\footnotesize \texttt{Travel forward past the wall with all the light switches and into the first room on your right.} 
\end{quote}
\textbf{Current observation $O_t$.} We use templates to convert the image caption $C_{t,j}$ and objects obtained $x_{t,j,0},\cdots, x_{t,j,M}$ from $I_{t,j}$ (\cref{sec:vision-text}).
For instance, if the agent is facing a heading of 90 degrees and an elevation of 0 degrees and there is a candidate navigable direction $I_{t,j}$ located at a heading of 120 degrees and an elevation of 0 degrees,  the text description for this view angle would be: \vspace{-0mm}
\begin{quote}
\footnotesize
\texttt{To your 30 degree right is ``\{$C_{t,j}$\}''. \\ Details: \{$x_{t,j,0}\}, \dots, \{x_{t,j,M}\}.$} \vspace{-0mm}
\end{quote}
We create such templates for all the navigable view angles $\{I_{t,0}, \dots, I_{t, V} \}$.
\vspace{-1mm}
\paragraph{Action $a_t$.} Selecting an action involves choosing a navigable view out of $O_t$ to move towards, i.e., $a_t \in O_t$. For example, suppose $a_t = I_{t, j}$, i.e., the agent decided to go to the $j$-th view angle. Then this is recorded as: \begin{quote} \vspace{-1mm}
\footnotesize
    \texttt{You go towards: ``$\{C_{t, j}\}$''}  \vspace{-1mm}
\end{quote}
To actually have the agent generate $a_t$ we simply decode from an LM's distribution, $p_{\text{LM}}(\cdot \, | \, D, U, H_t, O_t)$, via  greedy decoding. Here $H_t = \{O_i, a_i\}_{i=0}^{t-1}$ encodes the observation and action trajectory.\footnote{In general we found  the finetuned LM to have no issue  generating from the set of navigable directions (i.e., $\{C_{t,0}, \dots, C_{t,V} \}$) without constrained decoding.}
\paragraph{Updating trajectory history $H_t$.} 
We update the observation and action trajectory history via appending the text representations of $O_t$ and $a_t$ to $H_t$: \vspace{-1mm}
\begin{quote}
\footnotesize     \texttt{Step \{$t$\}: To your \{direction\_1\} is \{caption\_1\}; To your \{direction\_2\} is \{caption\_2\}; [...]; You chose: \{caption\_of\_selected\_direction\}}. \end{quote}
This history serves to inform the model about its current position within the high-level instruction, enabling it to make more informed decisions when selecting actions. 

\begin{figure*}
\centering
\includegraphics[width=0.9\linewidth]{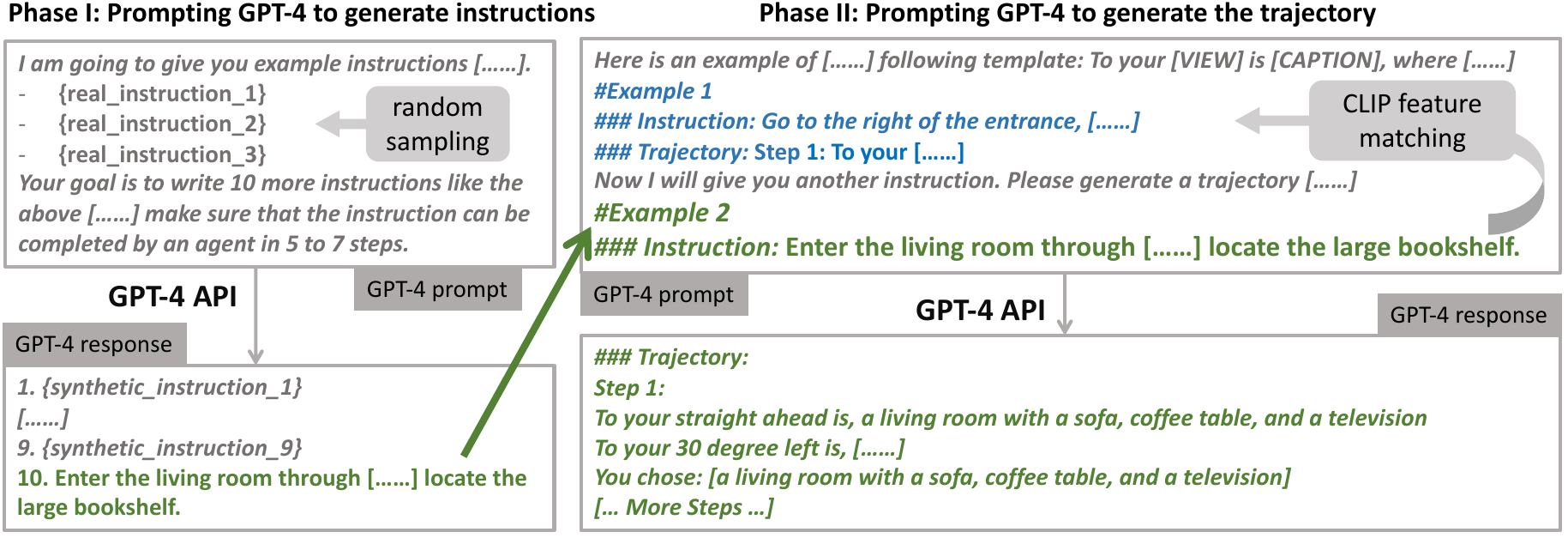}
\caption{Pipeline for generating synthetic navigation trajectories from GPT-4. We first prompt GPT-4 with 3 randomly sampled navigation instructions $U$ to generate 10 more synthetic navigation instructions (Phase 1). Then for each generated navigation instruction, we prompt GPT-4 to generate the trajectory that fulfills the generated instruction (Phase 2). See \cref{appx:temp} for details.}
\label{fig:pipeline}
\end{figure*}

\subsection{Imitation Learning on Demonstrations} \label{sec:bc}
 We create an instruction-following dataset by transforming the expert trajectory from the original dataset into instruction-following demonstrations. Formally, let $\mathcal{D} = \{ {W}^{(i)} \}_{i=1}^N$ be the set of training trajectories, where each $W^{(i)}$ can be represented as a natural language sequence from the above template, $W^{(i)} = (D^{(i)}, U^{(i)}, H_1^{(i)}, O_1^{(i)}, a_1^{(i)}, \dots, H_{T^{(i)}}^{(i)}, O_{T^{(i)}}^{(i)}, a_{T^{(i)}}^{(i)})$. Here $T^{(i)}$ is the number of actions in the example $W^{(i)}$, which is typically between 5 to 7.
Given the above, we optimize the log likelihood of the (language descriptions of) actions, i.e., the objective for trajectory $W^{(i)}$ is given by,
  $  \sum_{t=1}^{T^{(i)}} \log \, p_{\text{LM}}(a^{(i)}_t \, | \, D^{(i)}, U^{(i)}, H_t^{(i)}, O_t^{(i)})$.

While behavior cloning on gold trajectories is simple, it is prone to error propagation.  In particular, the history trajectory is obtained by a shortest-path algorithm (which has knowledge of the goal) and thus adheres closely to an optimal policy $\pi^*$. However, during prediction, trajectories can deviate significantly from the optimal policy, leading to a distribution shift that can adversely affect performance. To allow for the policy to recover from deviations from the optimal path, we adopt the following strategy to create our imitation learning dataset: (1) at each time step, we sample a random action with  probability $\rho$; (2) once a random action is selected, we use the shortest-path algorithm to obtain the ground truth next action; (3) we  repeat this process until the goal is reached; (4) once the goal is reached, this becomes part of the training demonstration data. (See \cref{app:rho} for details.)  

\begin{figure*}
\centering
\includegraphics[width=0.9\linewidth]{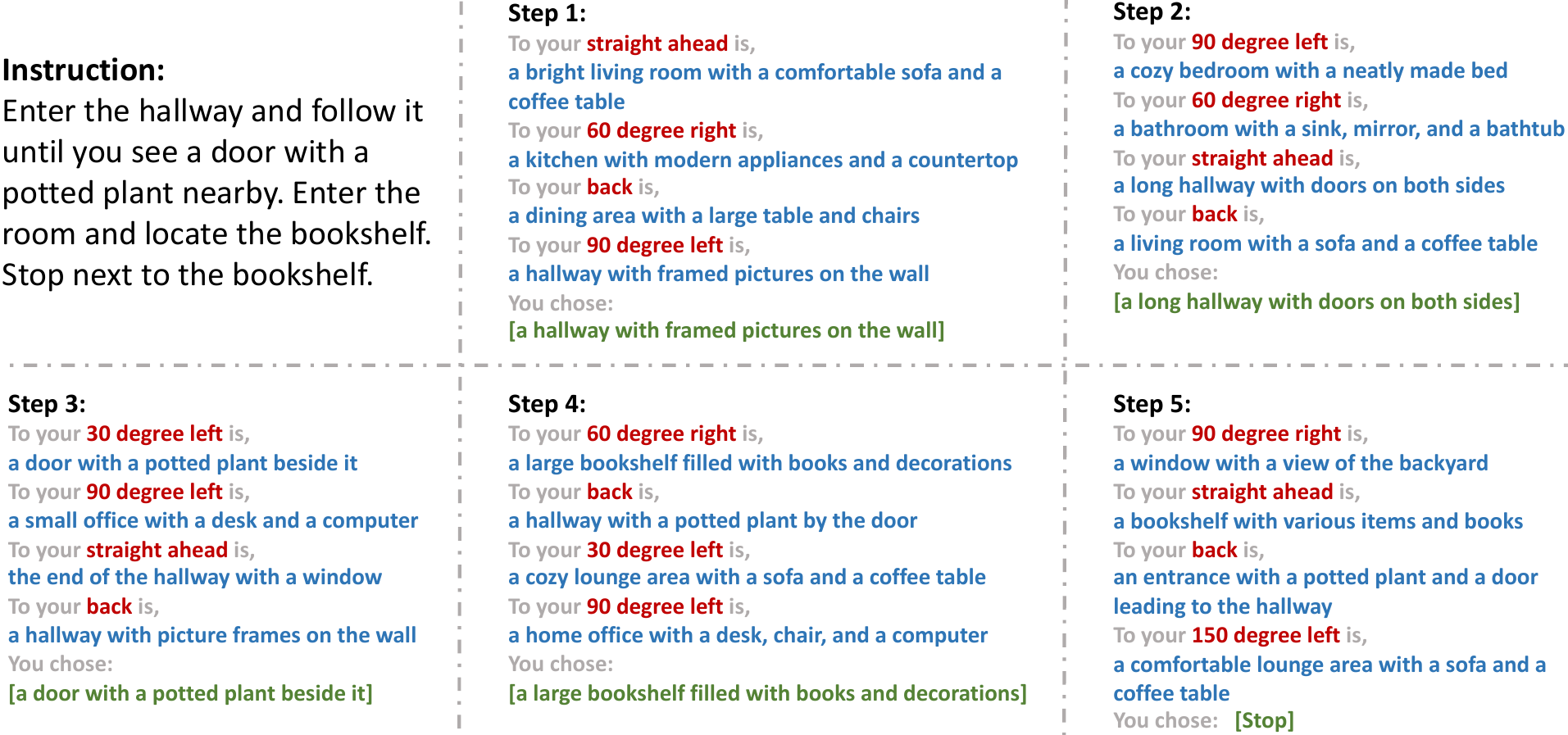}
\caption{An example of a generated trajectory from GPT-4. The example demonstrates a generated trajectory by following the pipeline in Figure~\ref{fig:pipeline}. See more examples in \cref{appx-gen}.}
\label{fig:spot}
\end{figure*}

\section{Empirical Study}\label{subsec:task}
Our primary experiments with \ours target the low-data setting, motivated by the observation that obtaining annotated data for embodied tasks such as vision-language navigation  can be very costly 
 (often more so than is the case for text-only or vision-only tasks). Specifically, we are interested in learning the most performant system based on a small number (10 or 100) of in-domain seed navigation trajectories.
 We sample our seed trajectories from the Room-to-Room (R2R) dataset~\citep{anderson2018vision}, a popular vision-and-language navigation dataset consisting of 21,567 navigation instructions in the Matterport3D environment. The dataset includes 90 scenes, with 61 scenes in the train and validation ``seen'' sets, and 11 scenes in the validation ``unseen'' set. Our 10-shot dataset is randomly sampled the train set within 1 scene, while our 100-shot dataset spans 2 scenes. 

\paragraph{Evaluation.} To contextualize our approach against prior work, we evaluate \ours on both  ``seen'' and ``unseen'' sets from R2R.  The ``seen'' set contains scenes identical to the training set (but the instructions and trajectories differ). However, this distinction is less important for our low-data regime, since we only make use of 
1 scene (for the 10-shot case) or 2 scenes (for the 100-shot case). I.e., the majority of scenes in the ``seen'' validation subset are actually never seen by the agent.

We use the standard R2R task performance metrics~\citep{anderson2018evaluation}: \textit{Navigation Error} (\texttt{NE}), the average distance between the agent's final position and the goal in meters; \textit{Success Rate} (\texttt{SR}), the ratio of trajectories in which the agent stopped within 3 meters of the goal; \textit{Oracle Success Rate} (\texttt{OSR}), the ratio of trajectories in which the agent stopped within 3 meters to the goal with a view of the goal; and \textit{Success} weighted by the normalized inverse of the \textit{Path Length} (\texttt{SPL}).

\subsection{Case Study 1: Language Enables Efficient Synthetic Data Generation}\label{sec:gpt4}

In NLP, obtaining synthetic data from an appropriately prompted large LM  with which to learn a smaller model has been shown to be an effective approach in data-scarce settings~\cite[][\textit{inter alia}]{wang2021want,lang2022co,alpaca,dai2023chataug,gunasekar2023textbooks}.\footnote{However see \citet{gudibande2023false} for a critical discussion of this approach.} However, this approach is difficult to extend to non-linguistic perceptual tasks such as VLN since generating realistic perception data is itself  difficult. In our first case study, we show that working in pure language space makes it possible to easily generate synthetic data from a large LM based on a few seed trajectories. We further show that finetuning a smaller LM on a mixture of synthetic and R2R trajectories improves upon vision-based models.

\paragraph{Synthetic trajectory generation.}
We generate synthetic trajectories by using only the 10 R2R trajectories from a single scene. In R2R each trajectory has 3 navigation instructions given by 3 different annotators. Thus we have 30 navigation instructions $\{U^{(i)}\}_{i=1}^{30}$ in total. Our data generation pipeline can be divided into two phases. In phase 1, we randomly choose 3 R2R instructions as prompt examples and ask GPT-4 to create 10 more instructions similar to the examples, as shown in Figure~\ref{fig:pipeline}. 
In phase 2, for each generated instruction, we prompt GPT-4 to generate a trajectory to fulfill the instruction, conditioned on a real demonstration instruction and trajectory. The real trajectory is obtained by selecting the trajectory whose instruction is closest to the synthetic instruction based on the CLIP~\citep{radford2021learning} text features. See Figure~\ref{fig:pipeline} for an overview and appendix~\ref{appx:temp} for the prompts.\footnote{We cannot entirely rule out the possibility that the GPT-4 training set included the text instructions seen in R2R. However, while the text instructions may have been encountered, the trajectories were unlikely to have been encountered during pretraining since we used vision systems to obtain the captions/objects. Out of the 10,000 generated instructions, we did not find any instructions that were in the actual R2R dataset.} 

We present an illustrative example in Figure~\ref{fig:spot} to demonstrate some qualitative characteristics of generated trajectories.  We find that the generated trajectories have: 
{\it strong real-world priors}, i.e., they exhibit  adherence to real-world room-object and object-object correlations, as evident from descriptions like ``\texttt{a bathroom with a sink, mirror, [...]}''; {\it spatial consistency}, where the examples maintain spatial consistency within the generated trajectories---for instance, in \texttt{Step 4}, the generated position identifies the door with a potted plant, consistent with its position in \texttt{Step 3}; and
{\it rich descriptions}---the generated trajectories have descriptive captions and objects that do not only relate to the given instruction, which makes it possible to successfully navigate through  language only.

\begin{table*}[t!]
\footnotesize
    \centering 
    \begin{adjustbox}{width=0.9\textwidth,center}
    \begin{tabular}{llHccccHcccc}
        \toprule 
    \multirow{1}{*}{\bf Methods} & {\bf\# real}
    & \multicolumn{5}{c}{{\bf Val Seen}} & \multicolumn{5}{c}{{\bf Val Unseen}} \\ 
    \rowcolor{white} & & TL & NE$\downarrow$ & OSR$\uparrow$ & SR$\uparrow$ & SPL$\uparrow$ & TL & NE$\downarrow$ & OSR$\uparrow$ & SR$\uparrow$ & SPL$\uparrow$ \\
    \midrule 
        Random Walk & 0 & 7.9 & 10.2 & 5 & 3 & 1 & 7.6 & 9.5 & 6 & 3 & 2\\
        LLaMA2-7B (Zero-shot) & 0 &  & 10.2 & 0 & 0 & 0 & & 9.5 & 0 & 0 & 0 \\
        GPT-4 (Zero-shot) & 0 & 7.9 & 10.5 & 15 & 9 & 8 & 8.1 & 10.2 & 17 & 10 & 8\\
        GPT-4 (Few-shot) & 1 & 9.2 & 10.1 & 17 & 10 & 9 & 9.8 & 9.9 & 22 & 13 & 11 \\
        NavGPT \citep{zhou2023navgpt} & 0 & - & - & - & - & - & 11.5 & 6.5 & 42 & 34 & 29\\
    \midrule
        RecBert \citep{Hong_2021_CVPR} & 10 & 6.1 & 10.8 & 9 & 7 & 6 & 6.6 & 10.1 & 13 & 9 & 9 \\
        DuET \citep{chen2022think} & 10 & 7.9 & 10.0 & 21 & 14  & 12 & 8.0 & 9.9 & 20 & 12  & 11 \\
         LLaMA2-7B & 10 & 8.0 & 10.2 & 15 & 11 & 10 & 7.8 & 9.6 & 16 & 11 & 9  \\
        \rowcolor{white}LangNav (with LLaMA2-7B) & 10 & 11.1 & \bf 7.5 & \bf 39 & \bf 31 & \bf 27 & 10.7 & \bf 7.0 & \bf 42 & \bf 32  & \bf 28  \\
    \midrule
        RecBert  \citep{Hong_2021_CVPR} & 100 & 8.7 & 9.3 & 27 & 20 &	19 & 8.7 & 9.4 & 26 & 19 & 17 \\
        DuET \citep{chen2022think}  & 100 & 10.2 & 9.2 & 31 & 21 & 18 &  10.1 & 9.4 & 32 & 23 & 19 \\
        LLaMA2-7B & 100 & 10.5 & 9.6 & 29 & 21 & 18 & 10.9 & 9.1 & 30 & 19 & 17  \\
        \rowcolor{white}LangNav (with LLaMA2-7B) & 100 & 10.8 & \bf 7.4 & \bf 40 & \bf 32 & \bf 28 & 10.6 & \bf 7.1 & \bf 45 & \bf 34 & \bf 29\\ 
    \bottomrule 
    \end{tabular}
    \end{adjustbox}
    \caption{Results on the R2R dataset with 10 or 100 real world trajectories. \ours finetunes  LLaMA2-7B  on the mixture of the real-world trajectories and 10,000 synthetic trajectories from GPT-4.}
    \label{tab:gpt4_result}
\end{table*}

\begin{table*}[tb!]
\footnotesize
    \centering 
    \begin{adjustbox}{width=0.9\textwidth,center}
    \begin{tabular}{lllllllccccccc}
    \toprule 
        \bf \# synthetic data & \bf Data-generating LM & \bf \# seed scenes &  NE$\downarrow$ & OSR$\uparrow$ & SR$\uparrow$ & SPL$\uparrow$  \\
    \midrule

2,000 & \small{GPT-3.5} & 10 & 9.8 & 31.0 &  15.6 & 12.2 \\    
2,000 & \small{GPT-4-turbo} & 1000 & 8.1 & 42.9 & 24.9 & 19.6 \\
\midrule
500& \small{GPT-4} & 10 & 8.0 & 38.2 & 24.5 & 20.6 \\
2,000 & \small{GPT-4} & 10 &  7.0 & 42.2 & 31.1 & 26.6\\
10,000 & \small{GPT-4} & 10 & 7.0 & 41.9 & 31.6 & 27.5 \\
2,000 + 2,000 & \small{GPT-4 + GPT-4-turbo} & 10 + 1000 & 7.1 & 43.2 & 32.6 & 28.3 \\
    
    \bottomrule 
    \end{tabular}
    \end{adjustbox}
    \caption{Performance on the R2R val unseen set as we vary the number of synthetically generated data, the underlying LM from which the synthetic data is generated, and number of seed scenes. Here the seed scenes refer to the scans from which trajectories are sampled, with multiple trajectories originating from each seed scene.}
    \label{tab:ablations}
\end{table*}

\paragraph{Experimental setup.} We compare \ours, which is a LLaMA2-7b model finetuned on a mixture of the 10,000 synthetic trajectories and 10/100 real trajectories,  against the following baselines: \textcolor{black}{\it 1. Random walk}, which selects a random action at each time step; \textcolor{black}{\it 2. GPT-4 (Zero-shot / Few-shot)}, where we prompt GPT-4 to complete the trajectory by changing the task description of the template in \cref{sec:prompt} (see \cref{appx-fs} for the full prompt).   For the few-shot baseline, due to the context length we use one full navigation trajectory as a demonstration example; \textcolor{black}{\it 3. NavGPT}, a recent work that also uses language as a perceptual representation (via image captioning and object detection) to perform navigation, but purely with GPT-4 \citep{zhou2023navgpt};  
\textcolor{black}{\it 4. RecBert}, a vision-based method that adopts a recurrent architecture proposed by \citet{Hong_2021_CVPR} to keep track of the trajectory history; \textcolor{black}{\it 5. DuET}, another vision-based method which additionally builds representations of the global map during learning \citep{chen2022think}; and \textcolor{black}{\it 6. LLaMA2-7B}, a language-only baseline that does not make use of the synthetic data from GPT-4. 

All finetuning methods use the same set of 10/100 trajectories. For these experiments, we did not find significant differences in performance when using the object detection module, and hence we only relied on the image captioning system to give the language description of each view angle in the prompt template. See \cref{appx:imp} for the  training setup including hyperparameters. 

\paragraph{Results.} The results are shown in \cref{tab:gpt4_result}. We find that our GPT-4 zero- and few-shot results underperform the NavGPT baseline despite using the same backbone model, potentially due to NavGPT's use of ground truth distance information and chain-of-thought prompting \citep{wei2022cot,kojima2023large}. Just finetuning LLaMA2-7B on the 10/100 gold trajectories does not perform well, although it is comparable to the vision-based policies. Training on a mixture of synthetic and R2R trajectories improves performance by a nontrivial margin, and the LLaMA2-7B-based \ours approaches the performance of NavGPT despite being many times smaller, indicating the effectiveness of our pipelined prompting strategy  for distilling the rich navigation-relevant world knowledge within GPT-4 to a smaller (and more efficient) language model.\footnote{While we still underperform NavGPT,  the performance gap is relatively narrow---within 1\% in terms of SPL. We observe that NavGPT employs object information filtered by a ground-truth depth map, limiting the data to objects within a 3-meter range. Such filtering is important to mitigate the redundancy and noise often associated with unfiltered object information (i.e., often too many irrelevant objects are detected). As highlighted in the NavGPT paper, this selective use of object information is important for achieving  good performance.}

\begin{table*}[tb!]
    \centering 
    \begin{adjustbox}{width=0.9\textwidth,center}
    \begin{tabular}{lllccccccccc}
        \toprule 
    \multirow{2}{*}{\bf Methods} & \multirow{1}{*}{\bf Pretraining} & \multirow{1}{*}{\bf R2R} &
     \multicolumn{4}{c}{{\bf Val Seen}} & \multicolumn{4}{c}{{\bf Val Unseen}}  \\
    & {\bf Data} & {\bf data } & NE$\downarrow$ & OSR$\uparrow$ & SR$\uparrow$ & SPL$\uparrow$ & NE$\downarrow$ & OSR$\uparrow$ & SR$\uparrow$ & SPL$\uparrow$ \\
    \midrule 
         & \multirow{2}{*}{R2R} & 10 & 10.8 & 9 & 7 & 6 & 10.1 & 13 & 9 & 9  \\
          & & 100 & 9.3 & 27 &20 & 19 & 9.4 & 26 & 19 & 17  \\
          \cline{2-11}
        \rowcolor{white}\cellcolor{white} &  & 0 & 9.5 & 12 & 8 & 4 & 9.0 & 12 & 7 & 3 \\        
        \rowcolor{white}\cellcolor{white} &  & 10 & 10.8 & 11 & 7 & 6 & 10.7 & 13 & 9 & 7 \\        
        \rowcolor{white}\cellcolor{white}\multirow{-5}{*}{RecBert} & \multirow{-3}{*}{ALFRED} & 100 & 9.9 & 22 & 18 & 17 & 10.2 & 23 & 15 & 14 \\
    \midrule
         & \multirow{2}{*}{None} & 10 & 10.3 & 17 & 10 & 8 & 9.8 & 20 & 11  & 8 \\
         & & 100 & 9.0 & 25 & 20  & 18 & 9.2 & 25 & 17 & 15 \\
         \cline{2-11}
       \rowcolor{white}\cellcolor{white} &  & 0 & 9.2 & 20 & 17 & 15 & 8.9 & 24 & 18 & 16 \\
       \rowcolor{white}\cellcolor{white} &  & 10 & 8.7 & 20 & 19 & 18 & 8.3 & 21 & 18 & 17 \\
       \rowcolor{white}\cellcolor{white}\multirow{-5}{*}{LangNav} & \multirow{-3}{*}{ALFRED} & 100 & 8.1 & 29 & 25 & 24 & 8.0 & 29 & 24 & 22\\
    \bottomrule 
    \end{tabular}
    \end{adjustbox}
    \caption{Domain transfer results where we pretrain a navigation agent on the simulated ALFRED environment (which uses rendered images) and finetune on the real-world R2R environment. We use LLaMA-7B~\citep{touvron2023llama} as our backbone model, and compare against the RecBert \citep{Hong_2021_CVPR} baseline. }
    \label{tab:sim2real}
\end{table*}

\paragraph{Ablation study.} In \cref{tab:ablations} we vary both the number of synthetic trajectories and the data-generating LM. Switching the synthetic data source from GPT-4 to GPT-3.5/GPT-4-turbo  results in noticeable declines, highlighting the importance of using a strong  LM. Increasing the number of synthetic trajectories increases performance, although the gains are marginal when going from  2,000 to 10,000 trajectories.  This is potentially due to the use of only 10 real trajectories from a single scene to prompt LLMs which results in lack of instruction diversity (see examples in \cref{app:instr-diversity}). To investigate the influence of the scene diversity, we use 1,000 navigation instructions sampled from various R2R scenes to prompt GPT-4-turbo\footnote{We chose GPT-4-turbo for its lower cost.} to generate 2,000 additional synthetic trajectories. We can see that although the 2,000 trajectories generated by GPT-4-turbo are not of the same quality as those generated by GPT-4, scaling up using these trajectories outperforms the results from the 10,000-trajectory set.
\subsection{Case Study 2: Language as a Bridge for Domain Transfer}\label{sec:sim}
We next experiment with using language as a domain-invariant representation space to transfer a policy that has been trained on a different (rendered) environment \citep[ALFRED;][]{ALFRED20}, to the real-world R2R environment. There are significant differences between ALFRED and R2R which makes straightforward domain transfer challenging. ALFRED uses images rendered from the synthetic AI2THOR environment~\citep{ai2thor}, while R2R, based on the Matterport3D, incorporates images captured from real indoor environments. ALFRED's navigation trajectories and instructions are also  simpler and shorter compared to R2R's instructions: R2R instructions involve guiding the agent between rooms, whereas ALFRED trajectories mainly keep the agent within a single room and provides instructions for household tasks. Finally in ALFRED, the agent is limited to rotating left/right by $90\degree$ and moving forward, while in R2R, the agent can move in any combination of 12 candidate heading directions and 3 elevation directions. See \cref{appx:gap} for detailed discussion of these differences, and see \cref{appx:imp} for the  experimental setup.

\paragraph{Results.}  We pretrain both RecBert~\citep{Hong_2021_CVPR}\footnote{Given that RecBert~\citep{Hong_2021_CVPR} has similar performance to DuET~\citep{chen2022think} in the few-shot setting according to \cref{tab:gpt4_result}, we choose RecBert to be the baseline because it is simpler and does not require a topological map.} and \ours on the simulated ALFRED environment and finetune on 0/10/100 R2R trajectories with object information. \ours uses LLaMA1-7b \citep{touvron2023llama} as the language model. The evaluation results for both methods are presented in \cref{tab:sim2real}. Interestingly, for RecBert,  pretraining on ALFRED actually \emph{hurts} performance, potentially due to the model's overfitting to the idiosyncracies of the rendered environment. And without any R2R data, RecBert performs at near chance, whereas \ours is able to exhibit some level of zero-shot transfer. Pretraining in ALFRED consistently leads to performance improvements for \ours.

\begin{figure*}
\centering
\includegraphics[width=0.95\linewidth]{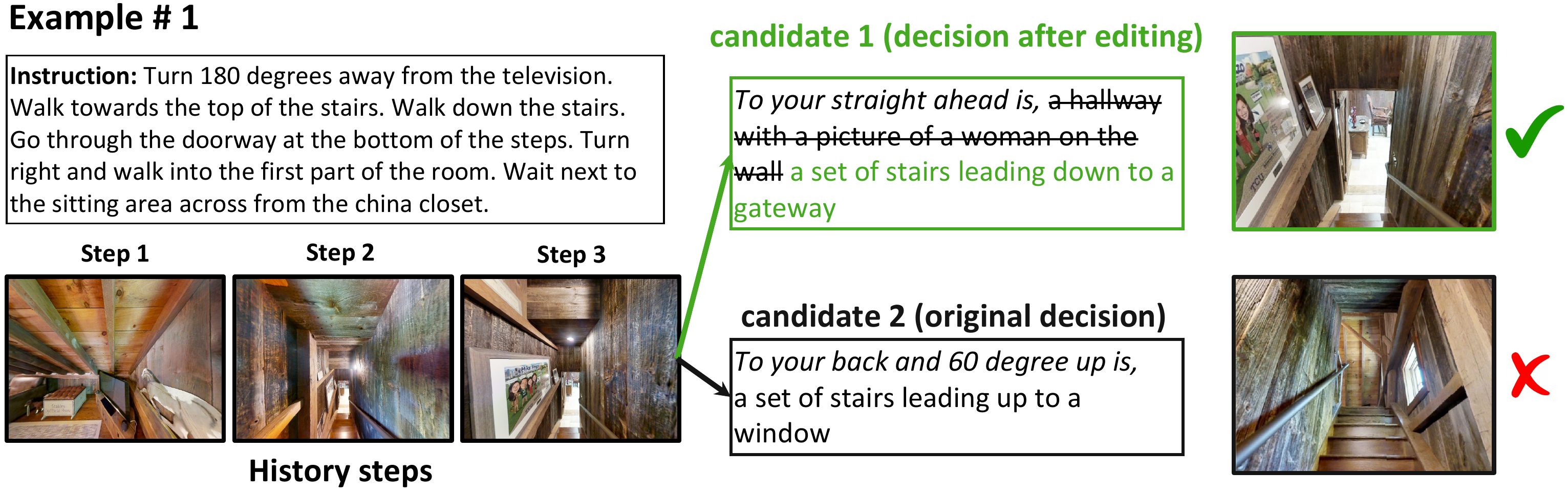}
\caption{Interpreting and editing a model's predictions through language. At the beginning, the agent incorrectly selected ``candidate 2'' to ascend the stairs. The failure might stem from the ambiguous interpretation of mistaking the stairs for a hallway in ``candidate 1''. After editing the description (marked in \textcolor{darkgreen}{green}), the agent correctly alters its choice to walk down the stairs.}
\label{fig:interpretability}
\end{figure*}

\subsection{Case Study 3: Combining Language and Vision Representations}
Our final case study explores whether language-based perceptual representations can improve performance \emph{on top of} traditional continuous vision features. This is motivated by the observation that (1) in the full data setting, \ours still underperforms the state-of-the-art  approaches which rely on pure vision features (see \cref{tab:comp_sota} of \cref{app:full-data}), and (2) realistic VLN scenarios would likely have access to continuous vision features as well. 

We extend the RecBert~\cite{Hong_2021_CVPR} by concatenating language features to the visual features to represent the candidate image view. Concretely, the original RecBert uses ResNet-152~\cite{he2016deep} to extract the visual feature to represent each view; our extension simply concatenates the caption representations (from BERT-base~\cite{devlin-etal-2019-bert}) to the image representation for each view. We train this new model on both the 100-shot and the full training set case. 

\paragraph{Results.} The results are listed in \cref{tab:lang_vis}. We find that  language features improve the performance in both 100-shot and full training set cases, which indicates that language as a perceptual representation can provide additional benefits on top of continuous visual features, even in non-low-data settings. This is potentially due to language serving as useful prior for aspects of images that are salient for navigation.
\begin{table}[tb!]
\centering

\begin{adjustbox}{width=0.94\columnwidth,center}
\begin{tabular}{llccc}
\toprule
\bf \# Training  & \bf Perceptual features & SR$\uparrow$ & SPL$\uparrow$ \\
\midrule
100 & Vision only & 19.0 & 17.4\\
100 & Vision + language & \text{19.3} & \text{18.0} \\
\midrule
Full train  & Vision only & 47.1 & 43.4 \\
Full train  & Vision + language & \text{48.8} & \text{44.1} \\
\bottomrule
\end{tabular}
\end{adjustbox}
\caption{Results when combining continuous visual features with language features with RecBert. Evaluations are conducted on R2R val unseen set.}
\label{tab:lang_vis}
\end{table}

\section{Discussion}\label{sec:discussion}

\paragraph{Interpretability and editability through language.} 
Our use of language as a ``bottleneck'' perceptual representation makes it possible to (more easily) \emph{interpret} and \emph{edit} a model's predictions. As a qualitative case study, we inspect trajectories where the model made a mistake and manually inspect the captions. We find that model mistakes are generally due to incorrect or ambiguous captions. We manually edit the captions to be correct, and find that in many cases, this is able to change the model's predictions to be correct. See Figure~\ref{fig:interpretability} for a concrete example.  We applied this procedure to 10 randomly selected trajectories which contained an error, and found that we were able to edit the model's decision to the correct one in 7 out of 10 trajectories. (For the other 3 trajectories, the failure was not due to incorrect captions). 

\paragraph{Disentangling vision and language models.} One the one hand, LangNav's use of a vision pipeline  might seem like a step back from pure deep learning-based approaches which generally favor learning everything ``end-to-end''. On the other, the disentangling of the image module from the language module means our approach can readily make use of independent advances in vision and language models. This might become especially important given the recent trend in only providing API access to state-of-the-art language models.

\paragraph{Non-standard navigation environments.} Our main experiments are on the R2R benchmark, which is realistic insofar as it makes use of real household environments. Another testbed for LangNav would be environments that lack existing datasets, such as offices or supermarkets. While the lack of existing benchmarks precludes our testing of LangNav on such non-standard environments, we performed a preliminary study where we tried generating synthetic trajectories from an office environment. We show an example in \cref{appx:exotic}, where we find that GPT-4 is able to generate synthetic trajectories that contain  common object-scene correlations in office environments and moreover exhibit great spatial consistency. Testing language as a perceptual representation in a variety of environments remains an interesting avenue for future work.
\section{Related Work}
\paragraph{Language Models for Task Planning.}
Several studies have explored language-based planning \citep{jansen2020visually, sharma2021skill, li2022pre, huang2022language, ahn2022can, huang2022inner}. \citet{huang2022language} use GPT-3 \citep{brown2020language} and Codex \citep{chen2021evaluating} for action plan generation with semantic translation using Sentence-RoBERTa \citep{huang2022language}. SayCan \citep{ahn2022can} grounds actions using FLAN \citep{wei2021finetuned} and action value functions \citep{shah2021value}. \citet{huang2022inner} explore incorporating grounded feedback into LLMs, while \citet{xiang2023language} propose enhancing LLMs' with embodied task instructions.

\paragraph{Instruction Tuning.}
 There has been much recent work finetuning smaller language models such as LLaMA on synthetic instruction-following data generated by GPT-3.5/GPT-4 \citep{peng2023instruction, alpaca, vicuna2023, LaMini-LM}. Existing works have generally focused on traditional language tasks. Our work instead finetunes LMs for embodied navigation tasks using language descriptions.

\paragraph{Vision-and-Language Navigation.}
There has been much work on vision and language navigation on the R2R dataset~\citep{anderson2018evaluation}. Approaches such as the speaker-follower model~\citep{fried2018speaker} and environmental dropout method \citep{tan2019learning}, reinforced cross-modal matching \citep{wang2019reinforced}, and self-monitoring \citep{ma2019self} have been proposed. Recent advancements include VLBERT-based methods \citep{Hong_2021_CVPR} and object-informed sequential BERT \citep{qi2021road}. \citet{qiao2022hop} incorporate additional pretext tasks into VLN pre-training based on \citet{Hong_2021_CVPR}. ALFRED~\citep{ALFRED20} involves interactive actions in a synthetic environment \citep{ai2thor}, with methods utilizing dense single vector representations \citep{ALFRED20, singh2021factorizing, pashevich2021episodic, kim2021agent, blukis2022persistent} or a panoramic view space \citep{suglia2021embodied}. CLIP-Nav \cite{dorbala2022clip} explores the zero-shot VLN with CLIP while \citet{kurita2020generative} proposes a generative language model-based navigation approach. For instruction synthesis, \citet{nguyen2019help} and \citet{thomason2020vision} studies rule-based instruction synthesis in Matterport3D environment. Finally, our work is closely related to \citet{zhou2023navgpt} and \citet{schumann2023velma}, which also use language descriptions of an agent's perceptual representation to perform navigation with an LM.

\section{Conclusion}

We show that we can learn to navigate in a real-world environments by using language as a perceptual representation. Language naturally abstracts away low-level perceptual details, which we find to be beneficial for
efficient data generation and sim-to-real transfer. However, this is also a serious limitation insofar as a picture really is worth a ``thousand words'' in some cases; we are certainly not suggesting the abandonment of traditional (continuous) vision features for vision-language navigation. But our case studies nonetheless demonstrate the promise of language as a perceptual representation for vision-language navigation.

\section*{Limitations}
While we find that LangNav is promising in settings where only a handful of real
trajectories are available, on the full dataset it still underperforms vision-based agents by a nontrivial margin, as shown in table~\ref{tab:comp_sota} of \cref{app:full-data}. This is especially true when compared to state-of-the-art approaches such as ScaleVLN \cite{wang2023scaling} which make use of large-scale pretraining data as well as more involved imitation/reinforcement learning algorithms that require access to an environment oracle. However, we note that while LangNav underperforms baselines in data-rich regimes, it overfits less to scenes seen during training, as demonstrated by the smaller drop in performance when applying the policy to unseen scenes during training. 

\section*{Acknowledgements}
This work is supported by MIT-IBM Watson AI Lab.

\bibliography{custom}

\appendix

\newpage

\begin{figure*}
\centering
\includegraphics[width=\linewidth]{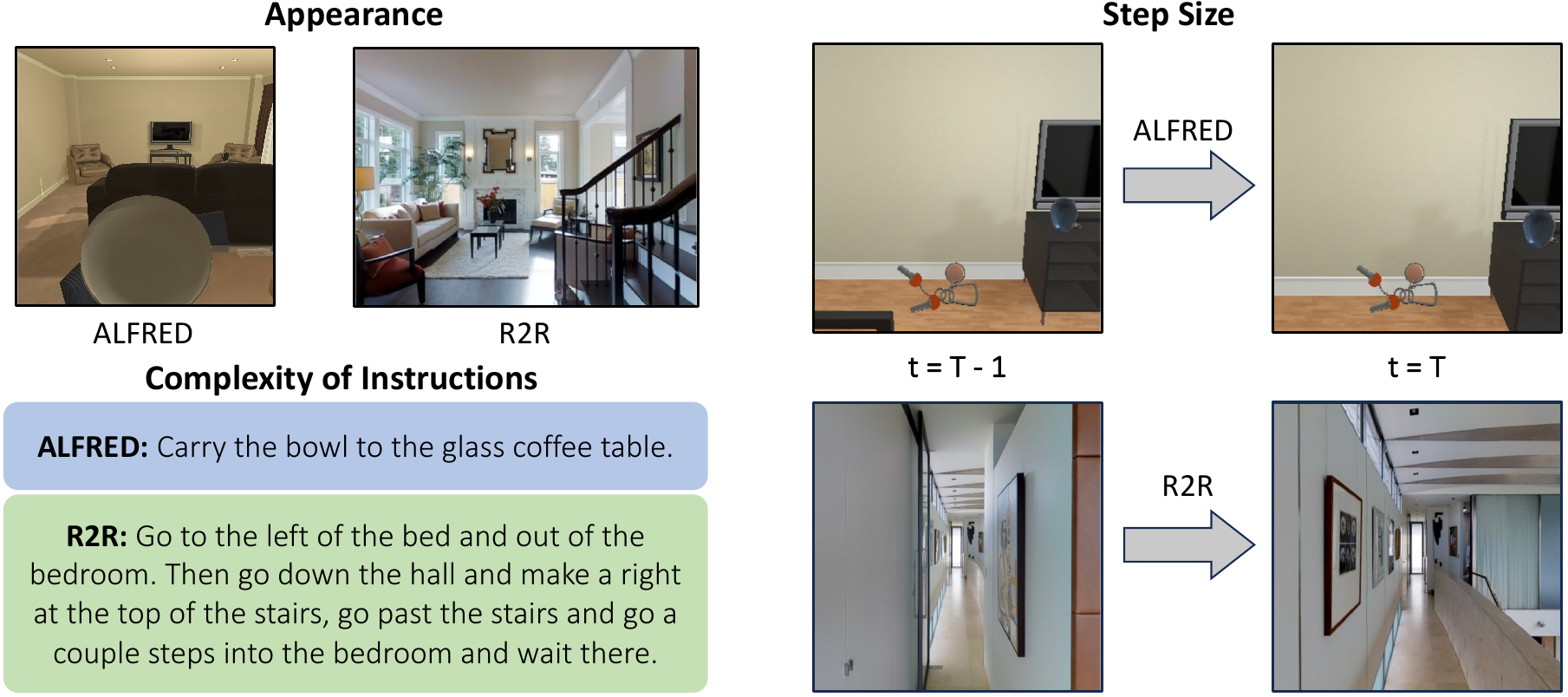}
\caption{Task gap between ALFRED and R2R. We highlight notable distinctions between the navigation tasks in ALFRED and R2R, encompassing variations in appearance, step size, and instruction complexity. See \cref{appx:gap} for more details.}
\vspace{-4mm}
\label{fig:gap}
\end{figure*}

\section{Implementations Details}\label{appx:imp}
We used the LLaMA-7B model~\citep{touvron2023llama} and the LLaMA2-7B model~\citep{touvron2023llama2} for our method, fine-tuning it on 72 V100-32GB GPUs with a batch size of 144. The training tokens had a maximum length of 1024, while during inference, the maximum length was set to 2048. The AdamW optimizer~\citep{loshchilov2017decoupled} with a learning rate of $2\times10^{-5}$ and weight decay of 0 was employed for optimization. The WarmupDecayLR learning rate scheduler was used for learning rate scheduling. For image captioning in both the R2R and ALFRED tasks, BLIP~\citep{li2022blip} was utilized. Deformable DETR~\citep{zhu2020deformable} was used for object detection in the R2R dataset, with suppression of outdoor object categories. We used the ground-truth object detection results provided in ALFRED when we generated the instruction-following pairs in \cref{sec:sim}. When prompting GPT-4 / GPT-4-turbo / GPT-3.5 API, we set the temperature as 1 and top\_p as 1. The cost of collecting the generated 10,000 trajectories by prompting GPT-4 API~\citep{openai2023gpt4} was around \$500. In the few-shot learning experiments in \cref{sec:gpt4} and \cref{sec:sim}, we set $\rho = 0$. While when fine-tuning with the full train set in \cref{appx:multitask}, we set $\rho = 0.2$. We pretrain on 128K ALFRED instruction-following pairs whose format is given in \cref{sec:prompt}. We augment the observations in ALFRED to 12 views and randomly mask a variable number of views to mimic the irregular number of candidates in R2R.
The RecBERT baselines in \cref{tab:gpt4_result}, \cref{tab:sim2real}, and \cref{tab:lang_vis} are pre-trained on 10/100 trajectories from R2R with masked language modeling (MLM) and single action prediction (SAP) tasks~\cite{hao2020prevalent}. The DUET baselines in \cref{tab:gpt4_result} are pre-trained on 10/100 trajectories with MLM, SAP, and masked region classification (MRC) tasks~\cite{chen2022think}.

\section{Differences between ALFRED and R2R.}\label{appx:gap}
The primary cause of the vast difference between ALFRED and R2R lies in their environmental rendering: ALFRED utilizes images from the synthetic AI2THOR environment~\citep{ai2thor}, whereas R2R, drawing from the Matterport3D database, features images from actual indoor environments. We summarize the differences in the following aspects:

\textbf{Visual appearance.} ALFRED uses images rendered from the synthetic AI2THOR environment, while R2R, based on the Matterport3D, incorporates images captured from real indoor environments. These image sources differ in texture, occlusion, illumination, and other visual aspects.

\textbf{Step size.}
There is a  difference in step sizes between the two tasks (see the right part of \cref{fig:gap}). ALFRED uses a step size of $0.25$ meters, while R2R has larger and more variable step sizes. To bridge this gap, we consolidate four consecutive \texttt{MoveAhead} steps into a single step along the ALFRED trajectory.

\textbf{Action type.}
A complete ALFRED trajectory includes not only navigation actions but also interaction actions, where the interaction actions are combined with a target object to change the state of the surrounding environment. In order to filter the interaction actions in ALFRED, we divide each ALFRED trajectory into multiple sub-trajectories and keep the sub-trajectories that are labeled with the \texttt{GotoLocation} tag.

\textbf{Instruction complexity.} Due to trajectory splitting, ALFRED's navigation trajectories and instructions appear simpler and shorter compared to R2R's instructions. R2R instructions involve guiding the agent between rooms, whereas ALFRED trajectories mainly keep the agent within a single room.

\textbf{Action space.} In ALFRED, the agent is limited to rotating left/right by $90\degree$ and moving forward, while in R2R, the agent can move in any combination of 12 candidate heading directions and 3 elevation directions. The number of available movement directions is irregular. This difference in action space makes R2R trajectories more human-like. To address this, we introduce randomness by adding or reducing a heading offset of $\pm 30\degree$ to the agent's direction at each step in ALFRED, allowing rotations of $30\degree$ or $60\degree$ in addition to $90\degree$.

\section{Performance on full data}
\label{app:full-data}
In Table~\ref{tab:comp_sota} we show the performance of LangNav on the full dataset, as well as comparisons against the state-of-the-art. While we find that LangNav is promising in settings where only a handful of real trajectories are available, on the full dataset it still underperforms vision-based agents by a nontrivial margin. This is especially true when compared to state-of-the-art approaches
such as ScaleVLN \cite{wang2023scaling} which make use of large-scale pretraining data as well as more
involved imitation/reinforcement learning algorithms that require access to an environment oracle
during training. However, we note that while LangNav underperforms baselines in data-rich regimes,
it overfits less to scenes seen during training, as demonstrated by the smaller drop in performance
when applying the policy to unseen scenes during training.

\section{Multi-Task Performance}\label{appx:multitask}
One of the advantages of our approach is its inherent suitability for multitasking. Similar to LLMs use instruction to handle multiple language tasks concurrently, we consolidate task information and inputs into instructions. To validate the multitasking capability of our method, we extend its application to the ALFRED task.

\begin{table*}[tb!]
    \centering 
    \footnotesize
    \begin{adjustbox}{width=\textwidth,center}
    \begin{tabular}{Hlllccccccc}
\toprule 
\bf Mem & \bf Method & \bf Training data & \bf Needs Oracle  & \bf Val Seen & \bf Val Unseen & \textbf{Drop} \\ 

\midrule

\multirow{4}{*}{Rec} & Seq2Seq (SF)~\cite{anderson2018vision} & R2R & No  & 38.6 & 21.8 & 16.8 \\

& RCM~\citep{wang2019reinforced} & R2R & Yes &  67.4 &  42.5 & 24.9  \\

& Speaker-Follower~\citep{fried2018speaker} & R2R+SpeakerAug. & Yes & 70.1 & 54.6 & 15.5 \\

& RecBert$^\dag$~\citep{Hong_2021_CVPR} & R2R+PREV & Yes & 71.8 & 54.5 & 17.3 \\

Seq & HAMT~\citep{chen2021history} & R2R+PREV & Yes& 75.0 & 65.7 & 9.3\\
\multirow{4}{*}{Map} & ScaleVLN~\cite{wang2023scaling} & R2R+PREV & No &  67.2 & 47.4  & 19.8 \\
 & ScaleVLN~\citep{wang2023scaling} & R2R+PREV & Yes &   76.9 & 72.9  & 4.0 \\
 & ScaleVLN~\citep{wang2023scaling} & R2R+PREV+ScaleVLN & No &  71.1 & 57.0  & 14.1 \\
 & ScaleVLN~\citep{wang2023scaling} & R2R+PREV+ScaleVLN & Yes &  80.5 & 78.1 & 2.4 \\
\midrule

\multirow{2}{*}{Lang} & LangNav & R2R & No & 55.0 & 43.2 &  11.8 \\

& LangNav (M) & R2R+ALFRED & No & 55.9 & 45.6 & 10.3 \\
\bottomrule
\end{tabular}
\end{adjustbox}
\vspace{-2mm}
\caption{Comparison with state-of-the-art vision-based methods on the R2R dataset when trained on the full dataset. We use  success rate (\texttt{SR}) as the performance metric. ``Needs oracle'' indicates that the model needs to rely on an oracle during training that can give the ground-truth next action based on a sampled path from the model.(M): Multi-Task model.}
\label{tab:comp_sota}
\end{table*}

\begin{table}[tb!]
\centering
\captionof{table}{Performance of the Multi-task Model on R2R. We demonstrate the multi-task capability of the LM agent. For single-task models, each model is trained within the task data. We trained the multi-task model with data from both R2R and ALFRED tasks.}
\begin{adjustbox}{width=\columnwidth,center}
\begin{tabular}{lcccc}
\toprule
\multirow{2}{*}{\bf Models} &  \multicolumn{2}{c}{\bf R2R Seen} & \multicolumn{2}{c}{\bf R2R Unseen} \\ 

 & SR$\uparrow$ & SPL$\uparrow$ & SR$\uparrow$ & SPL$\uparrow$\\ 
\midrule
 Single-Task &  55.0 & 51.0 & 43.2 & 37.9\\

 Multi-Task & \textbf{55.9} & \textbf{51.7} &\textbf{45.6} & \textbf{40.0}\\
\bottomrule
\end{tabular}
\end{adjustbox}
\label{tab:multi-r2r}
\end{table}

\begin{table}[tb!]
\caption{Performance of the Multi-task Model on ALFRED. ST: Single-Task. MT: Multi-Task.}
\centering
\begin{adjustbox}{width=\columnwidth,center}
\begin{tabular}{lcccc}
\toprule
\multirow{2}{*}{} & \multicolumn{2}{c}{\bf ALFRED Seen} & \multicolumn{2}{c}{\bf ALFRED Unseen} \\  
 & Task$\uparrow$ & GC$\uparrow$ & Task$\uparrow$ & GC$\uparrow$ \\ 
\midrule
 ST & 0.0 (0.0) & 6.0 (4.7) & 0.5 (0.1) & 9.5(7.8) \\
 MT & 0.0 (0.0) & \textbf{6.4 (5.0)} & \textbf{0.6 (0.2)} & \textbf{9.8 (7.8)} \\
\bottomrule
\end{tabular}
\end{adjustbox}
\label{tab:multi-alfred}
\end{table}

\paragraph{Metrics on ALFRED.}
We evaluate our model on ALFRED using two metrics: \textit{Task Success} (\texttt{Task}) and \textit{Goal-Condition Success} (\texttt{GC}). \texttt{Task Success} measures the ratio of trajectories where object positions and state changes accurately match all task goal conditions at the end. \texttt{GC} assesses the ratio of completed goal conditions in each action sequence. \texttt{Task Success} is only considered successful when \texttt{GC} is also $1$. On average, each ALFRED task has $2.55$ goal conditions. We also calculate the \textit{Path Length Weighted Metrics} (\texttt{PLW}) for both \texttt{Task} and \texttt{GC}, which normalize the metrics based on the actual action sequence length.

\paragraph{Results of the Multi-Task Model.}
 In ALFRED task, we set $\rho = 0$ as the expert policy in ALFRED is suboptimal. To save training time and balance the data amount between R2R and ALFRED, we utilize only 50\% of the training dataset, resulting in a dataset for ALFRED with 386K data pairs. For R2R task training, we maintain $\rho = 0.2$ and run each demonstration trajectory twice, resulting in a training set size of 235K for R2R. Consequently, the merged dataset for the multitask model contains a total of 621K instruction-following data pairs. We select VLN Bert~\citep{Hong_2021_CVPR} as the baseline for the R2R task and Seq2seq model~\citep{ALFRED20} for the ALFRED task. Given the substantial differences between the R2R task and the ALFRED task (\cref{sec:sim}), our method is, to the best of our knowledge, the first model that simultaneously addresses these two tasks. In \cref{tab:multi-r2r} and \cref{tab:multi-alfred}, we find that the multitask model exhibits superior performance compared to the single-task models. These results underscore the capability of our method to effectively handle multiple highly diverse tasks.

\begin{figure*}
\centering
\includegraphics[width=\linewidth]{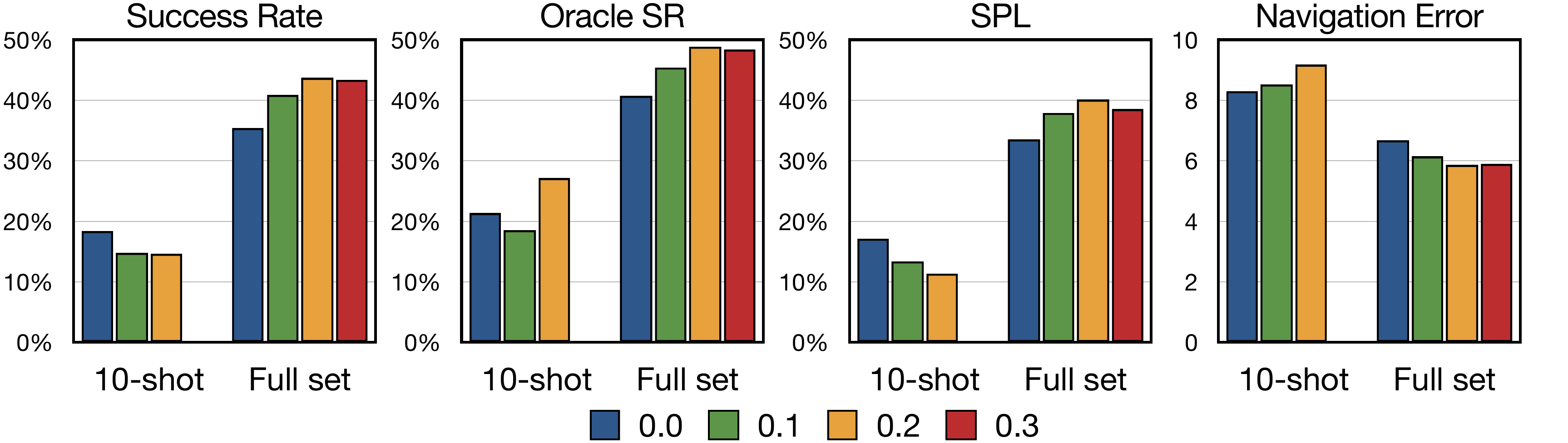}
\caption{Investigating the Impact of the Randomness Factor $\rho$ on Model Performance. This image caption depicts an ablation study exploring the influence of the randomness factor $\rho$ on our model's performance in both few-shot learning and full-set training scenarios. We test $\rho$ with values of 0.0, 0.1, 0.2, and 0.3.}
\label{fig:abl}
\end{figure*}

\section{Bias of generated navigation instructions}
\label{app:instr-diversity}
We list four generated instructions from one output of GPT-4 with 10 seed trajectories as we mentioned in \cref{sec:gpt4} as bellow:

\begin{prompt}[Examples of generated instructions]
    \texttt{1. Start from the main entrance door, pass the living room, and enter the kitchen on your right. Locate the refrigerator, then turn left and stop just before the dining table.} 

    \texttt{\\ 2. Navigate from the couch in the living room, move towards the mantel, and then stop next to the fireplace. Avoid any furniture and obstacles on your path.}
    
    \texttt{\\ 3. Begin at the foot of the bed in the master bedroom. Walk forward and enter the attached bathroom. Once you're inside, stop next to the bathtub.}
    
    \texttt{\\ 4. Start in the family room, walk towards the TV, then turn right and pass the bookshelf. Stop when you reach the large bay window overlooking the garden.}
\end{prompt}

We can see from the above synthetic instructions that (a) patterns of the synthetic instructions are similar, which are like "Start from place A, go pass place B, stop at place C", (b) scenes are limited to the living area and a single floor, however, the R2R tasks always require the agent navigating across floors and in some non-living area.

\section{Impact of the randomness factor}
\label{app:rho}
We conduct the ablation study to investigate the impact of the randomness factor $\rho$ on the model's performance in both few-shot learning and full-set fine-tuning scenarios. Interestingly, we observe different behaviors of the model with varying $\rho$ values in these scenarios. Fig.~\ref{fig:abl} illustrates our findings. In the 10-shot scenario, increasing $\rho$ negatively affected the model's performance. However, in the full-set scenario, the model achieved optimal performance at $\rho$ = 0.2.

\section{Example of the Full Trajectory}\label{sec:app-full}
We plot an example of a full text-based trajectory in R2R as we mentioned in \cref{sec:prompt} as bellow:

\begin{prompt}[Full R2R trajectory]

\begin{task_description}
You are a navigation agent who must navigate according to instructions given only descriptions of your current position via natural language. The natural language description is sometimes incorrect.
\end{task_description}

\begin{nav_instruction}
Go across the kitchen passed the pizza oven into the next room. Stop just inside the bedroom.  
\end{nav_instruction}

\begin{trajectory}

\textbf{Step 1:}
\begin{obs}
To your 60 degree left is,\\
a kitchen with a stove, sink, and refrigerator 
\\ 
Details:  oven, bowl, dining table, microwave, refrigerator 
\\
To your 30 degree left is,  \\ 
a kitchen with white cabinets and a white stove top oven \\ 
Details:  oven, bowl, microwave, potted plant, chair \\ 
To your 60 degree right is,  \\ 
a living room filled with furniture and a fire place \\ 
Details:  couch, chair \\  
To your back is,  \\ 
a room with a bed and a painting on the wall
\end{obs}

You chose:

\begin{ans}
a living room filled with furniture and a fire place
\end{ans}

\textbf{Step 2:}
\begin{obs}
To your 150 degree left is,  \\ a kitchen with a stove top oven sitting next to a counter \\ Details:  oven, dining table, microwave, refrigerator, chair \\  To your back is,  \\ a kitchen with a wooden table and white appliances \\ Details:  oven, bowl, dining table, microwave, refrigerator, apple, couch, chair \\  To your 30 degree right is,  \\ a living room with a couch a chair and a window \\ Details:  couch  \\ To your 120 degree right is,  \\ a dining room table with a bowl of fruit on it \\ Details:  chair, bowl, dining table \\  To your 150 degree right is,  \\ a bowl of fruit sits on a wooden table \\ Details:  couch, chair, bowl, dining table 
\end{obs}

You chose:

\begin{ans}
a living room with a couch a chair and a window
\end{ans}

\textbf{Step 3:}
\begin{obs}
    To your back is,  \\ a kitchen with a table, chairs, and stairs \\ Details:  oven, dining table, refrigerator, potted plant, chair \\ To your 150 degree right is,  \\ a room with a table, chairs, and stairs \\ Details:  oven, chair, refrigerator, dining table \\ To your straight ahead and 30 degree down is,  \\ a view of a hallway from the top of stairs \\ Details:  refrigerator \\ To your 90 degree right and 30 degree up is,  \\ a room with a staircase and a mirror on the wall \\ Details:  toilet
\end{obs}

You chose:

\begin{ans}
    a view of a hallway from the top of stairs
\end{ans}

\textbf{Step 4:}
\begin{obs}
    To your back and 60 degree up is,  \\ a living room filled with furniture and a ceiling fan \\ Details:  oven, potted plant, refrigerator \\  To your 90 degree left and 30 degree up is,  \\ a living room with a couch and a table \\ To your straight ahead and 30 degree up is,  \\ a bedroom with a bed and a mirror on the wall \\ Details:  bed
\end{obs}

You chose:

\begin{ans}
    a bedroom with a bed and a mirror on the wall
\end{ans}

\textbf{Step 5:}
\begin{obs}
    To your back is,  \\ a hallway leading to a kitchen and living room \\ Details:  refrigerator, potted plant  \\ To your 30 degree left is,  \\ a room with a wooden door and a mirror \\  To your straight ahead is,  \\ a bedroom with a bed, dresser, mirror and a ceiling fan \\ Details:  potted plant, bed \\ To your 30 degree right is,  \\ a bedroom with a bed and a ceiling fan \\ Details:  potted plant, bed \\  To your 60 degree right is,  \\ a bedroom with a bed, dresser and mirror \\ Details:  potted plant, bed
\end{obs}

You chose:
\begin{ans_nl}
    stop
\end{ans_nl}
\end{trajectory}
\end{prompt}

\section{Complete Prompt Template of Generating Trajectories for GPT-4}\label{appx:temp}
We list our complete templates for prompting GPT-4 to generate synthetic instructions (Phase I) and synthetic trajectories to fulfill the instruction (Phase II).

\begin{prompt}[Phase I: Prompt template for generating synthetic instructions]
\begin{systemmessage}
I am going to give you example instructions written by humans to train a deep learning-based navigation agent acting inside a home. These example instructions are intended to be completed by the navigation agent in 5-7 steps. 
\end{systemmessage}

\begin{fewshotexamples}

\texttt{- \{real\_instruction\_1\}}

\texttt{- \{real\_instruction\_2\}}

\texttt{- \{real\_instruction\_3\}}

\end{fewshotexamples}

\begin{usermessage}
Your goal is to write 10 more instructions like the above that can be used to train a navigation agent. Since the navigation agent will be navigating in different home environments, your instructions should also be diverse and cover a wide range of home environments and rooms. You should make sure that the instruction can be completed by an agent in 5 to 7 steps.
\end{usermessage}
\end{prompt}

\begin{prompt}[Phase II: Prompt template for generating synthetic trajectories]
\begin{systemmessage}
Here is an example of a large language model acting as a blind navigation agent in an indoor environment through text descriptions. The agent is given an instruction at the start and must follow the instruction. At each time step, the agent is given descriptions of its field of view via the following template:

\texttt{
\\
To your [VIEW] is [CAPTION]
\\ 
-  [VIEW] consists of the agent's visible field of view (e.g., 30 degrees right, 120 degrees left, etc.)
\\ 
- [CAPTION] is the text description of that view obtained from an image captioning model 
\\
}

\end{systemmessage}

\begin{fewshotexamples}
\# Example 1 
\texttt{
\\
\#\#\# Instruction: 
\\
\{real\_instruction\_example\} 
\\ 
\#\#\# Trajectory: 
\\
\{real\_trajectory\_example\}
\\
}
\end{fewshotexamples}

\begin{usermessage}
Now I will give you another instruction. Please generate a trajectory of 5-7 steps that would complete the instruction.

\# Example 2
\texttt{
\\
\#\#\# Instruction: 
\\
\{synthetic\_instruction\}
}
\end{usermessage}
\end{prompt}

\section{Prompts of Zero-shot and Few-shot Navigation for GPT-4}\label{appx-fs}
Here we attach the the task description $D$ in the prompt template for prompting GPT-4 to navigate in the R2R evaluation dataset.

\begin{prompt}[Zero-shot]
\begin{systemmessage}
You are a navigation agent who must navigate according to instructions given only descriptions of your current position via natural language. The natural language description is sometimes incorrect. 
\end{systemmessage}

\begin{usermessage}
At each step, you will be given several directions and captions for each direction. You must choose one direction by printing only the [caption\_of\_the\_direction] or choose "Stop" if you think the goal is reached.

For example:

Input:

\begin{verbatim}
To your [direction_1] is, [caption of the 
direction_1].
......
To your [direction_N] is, [caption of the 
direction_N].

You choose:

Output: [caption of the direction_3]
\end{verbatim}

Hint: You should use the information inside the instructions, history steps, and current observations to make the decision.
\end{usermessage}

\end{prompt}

\begin{prompt}[Few-shot]
\begin{systemmessage}
You are a navigation agent who must navigate according to instructions given only descriptions of your current position via natural language. The natural language description is sometimes incorrect. 
\end{systemmessage}

\begin{usermessage}
At each step, you will be given several directions and captions for each direction. You must choose one direction by printing only the [caption\_of\_the\_direction] or choose "Stop" if you think the goal is reached.

For example:

Input:


\begin{verbatim}
To your [direction_1] is, [caption of the 
direction_1].
......
To your [direction_N] is, [caption of the 
direction_N].

You choose:

Output: [caption of the direction_3]
\end{verbatim}

\end{usermessage}

\begin{fewshotexamples}

And here is an example trajectory:

\begin{verbatim}
### Instruction:
Go down the stairs. Turn right and go down 
the hallway. Turn right and stand near the 
fireplace.

### Trajectory:
Step 1:

To your straight ahead is,
an ornate doorway leading to another room

To your 60 degree right is,
a red carpeted staircase leading to a 
chandelier

To your 120 degree right is,
a room with a red carpet and a large mirror

To your back and 30 degree down is,
a room with a red carpet and two windows

To your 120 degree left is,
a room with a red carpet and gold trim

You chose:
a room with a red carpet and gold trim

Step 2:

To your 150 degree right is,
a very ornate staircase in a house with red 
and white striped chairs

To your back is,
a red carpeted hallway leading to a 
staircase

To your 150 degree left is,
a hallway with a red carpet and a chandelier

To your 120 degree left is,
a room with a red carpet and a chandelier

To your 90 degree left is,
a room with a chandelier and two windows

To your 60 degree left is,
a room with a red carpet and a large mirror

To your 30 degree right is,
a hallway with a red carpet and wooden doors

You chose:
a hallway with a red carpet and wooden doors

Step 3:

To your back is,
a hallway with a red carpet and a chandelier

To your straight ahead is,
a hallway with a red carpet and a gold 
ceiling

You chose:
a hallway with a red carpet and a gold 
ceiling

Step 4:

To your 90 degree right is,
a living room with a chandelier and a 
fireplace

To your 120 degree right is,
a room with a fireplace and a chandelier 
in it

To your back is,
a hallway with a red carpet and gold trim

To your 90 degree left is,
a room with a chandelier and a table in it

To your 30 degree right is,
a living room with a chandelier and a couch

You chose:
a living room with a chandelier and a 
fireplace

Step 5:

To your straight ahead is,
a living room filled with furniture and a 
fire place

To your 30 degree left is,
a living room filled with furniture and a 
chandelier

To your 150 degree right is,
a room with a chandelier and a table in it

To your back is,
a hallway with a chandelier and a mirror on 
the wall

To your 90 degree left is,
a living room filled with furniture and a 
chandelier

You chose:
Stop
\end{verbatim}

\end{fewshotexamples}

\begin{usermessage}
Hint: You should use the information inside the instructions, history steps, and current observations to make the decision.

Now let's start!
\end{usermessage}

\end{prompt}

\section{Examples of Generated Trajectories in Exotic Environments}\label{appx:exotic}
 To validate that LLMs can synthesize useful data in more exotic environments, we conduct an experiment where we handcraft a trajectory in a real office environment and then prompt GPT-4 to generate synthetic trajectories within the scope of the office environment. Here we first plot the language trajectory we have sampled from a real office environment. And then, we attach the generated trajectories from GPT-4 to demonstrate the ability of a strong LLM to generate trajectories in more exotic environments.
 
\begin{prompt}[Real-world trajectory]
\begin{nav_instruction_nl}
Start from the pantry on your left, pass the board room to enter the atrium. Go through the door to get into the office area. Turn right at the first corner, and then turn right to stop at the printer.
\end{nav_instruction_nl}
\begin{trajectory}
    \textbf{Step 1:}
    \begin{obs}
        To your straight ahead is,
        \\
        a long hallway with light on the ceiling and a fridge on the left
        \\
        To your 120 degree left is,
        \\
        a lounge with a screen on the wall, floor-to-ceiling winders, desks, and many high chairs
        \\ 
        To your 90 degree right is,
        \\
        an auditorium with three big screens and tiered seating facing the stage
    \end{obs}
    
    You chose:
    
    \begin{ans}
        a long hallway with light on the ceiling and a fridge on the left
    \end{ans}
    
    \textbf{Step 2:}
    \begin{obs}
        To your straight ahead is,
        \\
        a long hallway with four posters on the left leading to the atrium
        \\
        To your back is,
        \\
        a lounge with a sofa and a shelf stocked with snacks
    \end{obs}
    You chose:
    \begin{ans}
    a long hallway with four posters on the left leading to the atrium
    \end{ans}    

    \textbf{Step 3:}

    \begin{obs}
        To your straight ahead is,
        \\
        an atrium with lounge chairs on the right and a door on the straight ahead
        \\
        To your 90 degree left is,
        \\
        a front desk in the corner with a monitor on the desk.
        \\
        To your 90 degree right is,
        \\
        an exit gate towards a set of commercial elevators.
        \\
        To your back is,
        \\
        a long hallway with posters on the right wall 
    \end{obs}

    You chose:
    \begin{ans}
        an atrium with lounge chairs on the right and a door on the straight ahead
    \end{ans}

    \textbf{Step 4:}
    \begin{obs}
        To your straight ahead is,
        \\
        An office area where there is a black board and a work pod room.
        \\
        To your 90 degree left is,
        \\ 
        An array of office desks with monitors and office chairs.
        \\
        To your 90 degree right is,
        \\
        A large office area with multiple arrays of desks and chairs.
        \\
        To your back is,
        \\
        An exit door with a phone room on the right.
    \end{obs}

    You chose:
    \begin{ans}
        A large office area with multiple arrays of desks and chairs
    \end{ans}

    \textbf{Step 5:}
    \begin{obs}
        To you straight ahead is,
        \\
        A large office area with multiple arrays of desks and chairs, as well as trash bins.
        \\
        To your 60 degree left is,
        \\
        A black board in the middle of an office area.
        \\
        To your 90 degree right is,
        \\
        A copy/print room with recycle bins.
        \\
        To your back is,
        \\
        An array of office desks and chairs with a black board.
    \end{obs}

    You chose:
    \begin{ans}
        A copy/print room with recycle bins
    \end{ans}

    \textbf{Step 6:}
    \begin{obs}
        To your 90 degree right is,
        \\
        A printer on the ground and several instruction sheet on the wall
        \\
        To your back is,
        \\
        A black board and an array of office desks
    \end{obs}

    You chose:
    \begin{ans_nl}
        Stop
    \end{ans_nl}    
\end{trajectory}
\end{prompt}

\begin{prompt}[Synthetic trajectory \#1]
\begin{nav_instruction_nl}
Begin at the elevator lobby. Exit the lobby and turn left, heading towards the open office space. Walk past two rows of desks, then take a right. Continue straight until you see the coffee station. Turn left and stop at the second conference room.
\end{nav_instruction_nl}

\begin{trajectory}
    \textbf{Step 1:}
    \begin{obs}
        To your straight ahead is,
        \\
        an elevator lobby with multiple elevators and a directory stand
        \\
        To your 90 degree left is,
        \\
        a closed door with a sign "Stairs"
        \\
        To your 90 degree right is,
        \\
        a wide corridor leading to a reception area        
    \end{obs}

    You chose:
    \begin{ans}
        a wide corridor leading to a reception
    \end{ans}

    \textbf{Step 2:}
    \begin{obs}
        To your straight ahead is,
        \\
        a reception area with a white desk and three chairs
        \\
        To your 120 degree left is,
        \\
        a corridor leading to an open office space with several workstations
        \\
        To your 90 degree right is,
        \\
        a closed door marked "Emergency Exit"
    \end{obs}

    You chose:
    \begin{ans}
        a corridor leading to an open office space with several workstations
    \end{ans}

    \textbf{Step 3:}
    \begin{obs}
        To your straight ahead is,
        \\
        an open office space filled with rows of desks and office chairs
        \\
        To your 90 degree left is,
        \\
        an array of cubicles with computers.
        \\
        To your 90 degree right is,
        \\
        an exit door with a sign "Stairs"
    \end{obs}

    You chose:
    \begin{ans}
        an open office space filled with rows of desks and office chairs
    \end{ans}

    \textbf{Step 4:}
    \begin{obs}
        To your straight ahead is,
        \\
        an open office space with rows of desks and office chairs.
        \\
        To your back is,
        \\
        an office space with several workstations.
        \\
        To your 90 degree left is,
        \\
        a hallway leading to the break room.
        \\
        To your 90 degree right is,
        \\
        a corner office space with a large round table.
    \end{obs}

    You chose:
    \begin{ans}
        an office space with rows of desks and office chairs.
    \end{ans}

    \textbf{Step 5:}
    \begin{obs}
        To your straight ahead is,
        \\
        an open office space with rows of desks and office chairs.
        \\
        To your 30 degree left is,
        \\
        a meeting room with a large table and multiple chairs.
        \\
        To your 120 degree right is,
        \\
        a corridor leading to a door.
    \end{obs}

    You chose:
    \begin{ans}
        a corridor leading to a door.
    \end{ans}

    \textbf{Step 6:}
    \begin{obs}
        To your straight ahead is,
        \\
        a door at the end of the hallway
        \\
        To your 90 degree right is,
        \\
        a coffee station with a variety of coffee machines
    \end{obs}

    You chose:
    \begin{ans}
        a coffee station with a variety of coffee machines
    \end{ans}

    \textbf{Step 7:}
    \begin{obs}
        To your straight ahead is,
        \\
        a small kitchenette with a fridge and microwave
        \\
        To your immediate left is,
        \\ 
        a conference room featuring a large glass table
        \\
        To your second left is,
        \\
        a second conference room with a project screen and chairs
    \end{obs}

    You chose:
    \begin{ans}
        a second conference room with a project screen and chairs
    \end{ans}

    \textbf{Step 8:}
    \begin{obs}
        To your straight ahead is,
        \\
        a conference room with a projection screen and chairs.
        \\
        To your back is,
        \\
        a coffee station with various coffee machines.
    \end{obs}

    You chose:
    \begin{ans_nl}
        Stop
    \end{ans_nl}
    
\end{trajectory}
\end{prompt}

\begin{prompt}[Synthetic trajectory \#2]
\begin{nav_instruction_nl}
Start outside the IT department's office. Walk straight and take the first left towards the stairwell. Go up one flight of stairs. Upon reaching the top, turn right and proceed to the end of the corridor. The library is on the right.
\end{nav_instruction_nl}

\begin{trajectory}
    \textbf{Step 1:}
    \begin{obs}
        To your straight ahead is,
    \\
    the hallway leading to an intersection with doors for the office spaces on the sides
    \\
    To your 90 degree left is,
    \\
    a door with a sign “IT Department” followed by some tech carts
    \\
    To your 90 degree right is,
    \\
    a hallway towards the elevators with seating areas on the sides
    \end{obs}    

    You chose:
    \begin{ans}
        the hallway leading to an intersection with doors for the office spaces on the sides
    \end{ans}

    \textbf{Step 2:}
    \begin{obs}
        To your straight ahead is,
        \\
        a hallway with the elevator and murals on its sides
        \\
        To your 90 degree left is,
        \\
        a hallway that moves towards a stairwell
        \\
        To your back is,
        \\
        a hallway leading towards the IT department’s office        
    \end{obs}

    You chose:
    \begin{ans}
        a hallway that moves towards a stairwell        
    \end{ans}

    \textbf{Step 3:}
    \begin{obs}
        To your straight ahead is,
        \\
        a stairwell leading upwards with railing and stair lights
        \\
        To your 90 degree left is,
        \\
        a hallway towards a coffee machine and cafeteria entrance
        \\ 
        To your 90 degree right is,
        \\
        a door to a rest room with signage
        \\ 
        To your back is,
        \\
        a hallway with murals leading back to the IT department's office.
    \end{obs}

    You chose:
    \begin{ans}
        a stairwell leading upwards with railing and stair lights
    \end{ans}

    \textbf{Step 4:}
    \begin{obs}
        To your straight-ahead is,
        \\
        another flight of stairs leading to a large window
        \\
        To your 90 degree right is,
        \\
        a hallway lined with posters and a bench
        \\
        To your back is,
        \\
        a flight of stairs that leads down with railing and stair lights        
    \end{obs}

    You chose:
    \begin{ans}
        a hallway lined with posters and a bench
    \end{ans}

    \textbf{Step 5:}
    \begin{obs}
        To your straight-ahead is,
        \\
        a corridor leading to an intersection with a water fountain to the right
        \\
        To your 90 degree left is,
        \\
        a large window that illuminates the corridor and gives a view of the street
        \\
        To your 90 degree right is,
        \\
        a library entrance with a signage
        \\
        To your back is,
        \\
        the hallway lined with posters and a bench        
    \end{obs}

    You chose:
    \begin{ans}
        a library entrance with a signage        
    \end{ans}

    \textbf{Step 6:}
    \begin{obs}
        To your straight ahead is,
        \\
        a library filled with wooden bookshelves and desks with desk lamps
        \\
        To your back is,
        \\
        a corridor leading back to the stairwell
    \end{obs}

    You chose:
    \begin{ans_nl}
        Stop
    \end{ans_nl}
        
\end{trajectory}
\end{prompt}

\section{More Examples of Generated Trajectories}\label{appx-gen}
We plot more examples of the generated trajectories from GPT-4 below:

\begin{prompt}[Example \#1]
    \begin{nav_instruction_nl}
        Begin in the basement, walk towards the pool table. Pass the pool table and ascend the staircase. At the top of the stairs, turn right and walk through the hallway. Turn left into the den, walk through and stop by the window.
    \end{nav_instruction_nl}

    \begin{trajectory}
        \textbf{Step 1:}
        \begin{obs}
            To your straight ahead is, \\
            a basement with a pool table, a bar, and a seating area \\
            To your 90 degree left is, \\
            a wall with various framed photos and a door \\
            To your 150 degree right is, \\
            a storage area filled with boxes and shelves
        \end{obs}
        
        You chose:
        
        \begin{ans}
            a basement with a pool table, a bar, and a seating area
        \end{ans}
        
        \textbf{Step 2:}
        \begin{obs}
            To your back is, \\
            a basement entrance with a staircase leading down \\
            To your 120 degree left is, \\
            a basement with a bar, television, and several chairs \\
            To your straight ahead is, \\
            a basement with a pool table, a dartboard, and a bar \\
            To your 90 degree right is, \\
            a staircase leading upwards with a wooden railing
        \end{obs}
        
        You chose:
        
        \begin{ans}
            a basement with a pool table, a dartboard, and a bar
        \end{ans}
        
        \textbf{Step 3:}
        \begin{obs}
            To your back is, \\
            a basement with a seating area and framed photos on the wall \\
            To your straight ahead is, \\
            a pool table surrounded by chairs and a dartboard on the wall \\
            To your 150 degree left is, \\
            a basement with a bar, television, and several chairs \\
            To your 90 degree right is, \\
            a staircase leading upwards with a wooden railing
        \end{obs}
        
        You chose:
        
        \begin{ans}
            a staircase leading upwards with a wooden railing
        \end{ans}
        
        \textbf{Step 4:}
        \begin{obs}
            To your back is, \\
            a basement with a pool table and a dartboard on the wall \\
            To your 90 degree left is, \\
            a hallway with various framed photos on the walls \\
            To your straight ahead is, \\
            a den with a large bookshelf, couch, and coffee table
        \end{obs}
        
        You chose:
        
        \begin{ans}
            a hallway with various framed photos on the walls
        \end{ans}
        
        \textbf{Step 5:}
        \begin{obs}
            To your back is, \\
            a staircase leading downwards into the basement \\
            To your straight ahead is, \\
            a long hallway with several doors leading to other rooms \\
            To your 120 degree right is, \\
            a den with a comfortable seating area and a bookshelf
        \end{obs}
        
        You chose:
        
        \begin{ans}
            a long hallway with several doors leading to other rooms
        \end{ans}
        
        \textbf{Step 6:}
        \begin{obs}
            To your back is, \\
            a hallway with various framed photos on the walls \\
            To your 90 degree left is, \\
            a den filled with cozy furniture, bookshelves, and a television \\
            To your 90 degree right is, \\
            a bedroom with a large bed, dresser, and a window
        \end{obs}
        
        You chose:
        
        \begin{ans}
            a den filled with cozy furniture, bookshelves, and a television
        \end{ans}
        
        \textbf{Step 7:}
        \begin{obs}
            To your back is, \\
            a hallway leading to other parts of the house \\
            To your 90 degree left is, \\
            a den with several bookshelves, a fireplace, and a large rug \\
            To your straight ahead is, \\
            a window overlooking the garden outside
        \end{obs}
        
        You chose:
        
        \begin{ans_nl}
            Stop
        \end{ans_nl}
        
    \end{trajectory}
\end{prompt}

\begin{prompt}[Example \#2]
    \begin{nav_instruction_nl}
        From the garage entrance, walk into the kitchen. Turn left and walk past the stove. Continue straight into the family room, then walk towards the sliding glass doors. Step outside and stop near the barbecue grill.
    \end{nav_instruction_nl}

    \begin{trajectory}
        \textbf{Step 1:}
        \begin{obs}
            To your straight ahead is, \\
            a clean garage with a parked car and a workbench \\
            To your 90 degree left is, \\
            a laundry room with a washer and a dryer \\
            To your back is, \\
            a driveway leading up to a garage door \\
            To your 120 degree left is, \\
            a garage entrance with an open door leading inside the house
        \end{obs}
        
        You chose:
        
        \begin{ans}
            a garage entrance with an open door leading inside the house
        \end{ans}
        
        \textbf{Step 2:}
        \begin{obs}
            To your 30 degree left is, \\
            a kitchen with a stove, a refrigerator, and a microwave oven \\
            To your 90 degree right is, \\
            a small mudroom with a shoe rack and a coat hook \\
            To your back is, \\
            a view of a garage with a car and a workbench
        \end{obs}
        
        You chose:
        
        \begin{ans}
            a kitchen with a stove, a refrigerator, and a microwave oven
        \end{ans}
        
        \textbf{Step 3:}
        \begin{obs}
            To your straight ahead is, \\
            a large kitchen with a stove and a range hood \\
            To your 30 degree right is, \\
            a kitchen with a center island and a sink on the countertop \\
            To your 150 degree left is, \\
            a kitchen with a refrigerator and a microwave oven \\
            To your 120 degree right is, \\
            a tiled floor leading to a hallway with a picture on the wall
        \end{obs}
        
        You chose:
        
        \begin{ans}
            a large kitchen with a stove and a range hood
        \end{ans}
        
        \textbf{Step 4:}
        \begin{obs}
            To your straight ahead is, \\
            a cozy family room with a TV and a fireplace \\
            To your 90 degree right is, \\
            a kitchen with a sink and a dishwasher \\
            To your back is, \\
            a kitchen with a countertop, a cooktop, and an oven
        \end{obs}
        
        You chose:
        
        \begin{ans}
            a cozy family room with a TV and a fireplace
        \end{ans}
        
        \textbf{Step 5:}
        \begin{obs}
            To your 30 degree left is, \\
            a sliding glass door leading to the backyard \\
            To your 90 degree right is, \\
            a comfortable sofa with a coffee table in front of a TV \\
            To your back is, \\
            a large kitchen with a stove and a range hood
        \end{obs}
        
        You chose:
        
        \begin{ans}
            a sliding glass door leading to the backyard
        \end{ans}
        
        \textbf{Step 6:}
        \begin{obs}
            To your straight ahead is, \\
            a well-maintained backyard with a barbecue grill and a dining table \\
            To your 90 degree left is, \\
            a sliding glass door leading inside the family room \\
            To your back is, \\
            a cozy family room with a TV and a fireplace
        \end{obs}
        
        You chose:
        
        \begin{ans}
            a well-maintained backyard with a barbecue grill and a dining table
        \end{ans}
        
        \textbf{Step 7:}
        \begin{obs}
            To your 90 degree left is, \\
            a barbecue grill on a patio in the backyard \\
            To your straight ahead is, \\
            a backyard with a dining table and chairs \\
            To your 120 degree right is, \\
            a backyard with a garden and a swing set \\
            To your back is, \\
            a sliding glass door leading inside the family room
        \end{obs}
        
        You chose:
        
        \begin{ans_nl}
            Stop
        \end{ans_nl}
        
    \end{trajectory}

\end{prompt}

\begin{prompt}[Example \#3]
    \begin{nav_instruction_nl}
    Stand in front of the fireplace in the living room. Walk diagonally across the room, passing the sofa. Proceed through the door into the hallway. Turn left and continue toward the guest room. Stop just before entering and wait.
    \end{nav_instruction_nl}

    \begin{trajectory}
        \textbf{Step 1:}
        \begin{obs}
            To your straight ahead is, \\
            a living room with a fireplace and two chairs \\
            To your 30 degree left is, \\
            a living room with a large window and a dark brown couch \\
            To your 90 degree right is, \\
            a living room with a flat screen tv and a view of the backyard
        \end{obs}
        
        You chose:
        
        \begin{ans}
            a living room with a fireplace and two chairs
        \end{ans}
        
        \textbf{Step 2:}
        \begin{obs}
            To your 60 degree left is, \\
            a living room with a fireplace and a floor lamp \\
            To your 30 degree right is, \\
            a living room with a dark brown couch and a coffee table \\
            To your straight ahead is, \\
            a living room with a white rug in the middle
        \end{obs}
        
        You chose:
        
        \begin{ans}
            a living room with a dark brown couch and a coffee table
        \end{ans}
        
        \textbf{Step 3:}
        \begin{obs}
            To your back is, \\
            a living room with a fireplace and a christmas tree \\
            To your 150 degree left is, \\
            a living room with two chairs and a painting on the wall \\
            To your straight ahead is, \\
            a wooden door leading to a hallway
        \end{obs}
        
        You chose:
        
        \begin{ans}
            a wooden door leading to a hallway
        \end{ans}
        
        \textbf{Step 4:}
        \begin{obs}
            To your 90 degree left is, \\
            a hallway with a view of a staircase and a table \\
            To your straight ahead is, \\
            a hallway with a painting on the wall and an open door
        \end{obs}
        
        You chose:
        
        \begin{ans}
            a hallway with a painting on the wall and an open door
        \end{ans}
        
        \textbf{Step 5:}
        \begin{obs}
            To your back is, \\
            a hallway with a wooden floor and a closed door \\
            To your 120 degree left is, \\
            a guest bedroom with a neatly made bed and a dresser \\
            To your 30 degree right is, \\
            a hallway with white walls and floor-to-ceiling mirrors
        \end{obs}
        
        You chose:
        
        \begin{ans_nl}
            Stop just before entering the guest bedroom
        \end{ans_nl}
    \end{trajectory}
\end{prompt}

\end{document}